\documentclass{article}

\usepackage[preprint]{neurips_2024}
\usepackage{comment}

\usepackage[utf8]{inputenc} 
\usepackage[T1]{fontenc}    
\usepackage{hyperref}       
\usepackage{url}            
\usepackage{booktabs}       
\usepackage{amsfonts}       
\usepackage{nicefrac}       
\usepackage{microtype}      
\usepackage{xcolor} 
\usepackage{graphicx}
\usepackage{amsthm}
\theoremstyle{plain}
\newtheorem{theorem}{Theorem}[section]

\newtheorem{lemma}[theorem]{Lemma}

\theoremstyle{definition}

\theoremstyle{remark}
\newtheorem{remark}[theorem]{Remark}
\usepackage{bm}
\usepackage{amsmath}
\usepackage{amssymb}
\usepackage{mathtools}
\usepackage{algorithm}
\usepackage{algorithmic}
\usepackage[capitalize,noabbrev]{cleveref}

\usepackage{tablefootnote}
\usepackage{wrapfig}

\newcommand{\nl}[1]{{\color{red}#1}}

\usepackage{multirow}
\usepackage{adjustbox}
\usepackage{url}
\usepackage{booktabs}

\usepackage{caption}
\usepackage{subfigure}
\usepackage{color, colortbl}
\definecolor{Gray}{gray}{0.9}
\newcolumntype{g}{>{\columncolor{Gray}}c}

\newcommand{\real}{\mathbb{R}}

\newcommand{\mcD}{\mathcal{D}}
\newcommand{\mcE}{\mathcal{E}}

\newcommand{\mcL}{\mathcal{L}}

\usepackage{xcolor}
\usepackage{soul}

\def\omg{{\Omega}}

\def \fb{\bm{f}}

\def \Fb{\bm{F}}

\def \ub{\bm{u}}
\def \Ub{\bm{U}}

\def \xb{\bm{x}}

\def \Xb{\bm{X}}
\def \Wb{\bm{W}}

\newcommand{\vertii}[1]{{\left\vert\left\vert #1
    \right\vert\right\vert}}    
    
\newcommand{\verti}[1]{{\left\vert #1
    \right\vert}}

\title{Nonlocal Attention Operator: Materializing Hidden Knowledge Towards Interpretable Physics Discovery}

\author{%
  Yue Yu\thanks{Corresponding Author} \\
  Department of Mathematics, \\
  Lehigh University, \\
  Bethlehem, PA 18015, USA \\
  \texttt{yuy214@lehigh.edu} \\
  \And
  Ning Liu \\
  Global Engineering and \\Materials, Inc., \\
  Princeton, NJ 08540, USA \\
 \And
  Fei Lu \\
  Department of Mathematics, \\
  Johns Hopkins University, \\
  Baltimore, MD 21218, USA \\
  \And
  Tian Gao \\
  IBM Research, \\
  Yorktown Heights, \\NY 10598, USA \\
  \And
  Siavash Jafarzadeh \\
  Department of Mathematics, \\
  Lehigh University, \\
  Bethlehem, PA 18015, USA\\
  \And 
  Stewart Silling \\
  Center for Computing Research, \\
  Sandia National Laboratories, \\
  Albuquerque, NM 87123, USA\\
}

\begin{document}

\maketitle


\begin{abstract}
Despite the recent popularity of attention-based neural architectures in core AI fields like natural language processing (NLP) and computer vision (CV), their potential in modeling complex physical systems remains under-explored. Learning problems in physical systems are often characterized as discovering operators that map between function spaces based on a few instances of function pairs. This task frequently presents a severely ill-posed PDE inverse problem. In this work, we propose a novel neural operator architecture based on the attention mechanism, which we coin Nonlocal Attention Operator (NAO), and explore its capability towards developing a foundation physical model. In particular, we show that the attention mechanism is equivalent to a double integral operator that enables nonlocal interactions among spatial tokens, with a data-dependent kernel characterizing the inverse mapping from data to the hidden parameter field of the underlying operator. As such, the attention mechanism extracts global prior information from training data generated by multiple systems, and suggests the exploratory space in the form of a nonlinear kernel map. Consequently, NAO can address ill-posedness and rank deficiency in inverse PDE problems by encoding regularization and achieving generalizability. We empirically demonstrate the advantages of NAO over baseline neural models in terms of generalizability to unseen data resolutions and system states. Our work not only suggests a novel neural operator architecture for learning interpretable foundation models of physical systems, but also offers a new perspective towards understanding the attention mechanism.
\end{abstract}


\section{Introduction}


The interpretability of machine learning (ML) models has become increasingly important from the security and robustness standpoints \citep{rudin2022interpretable,molnar2020interpretable}. This is particularly true in physics modeling problems that can affect human lives, where not only the accuracy but also the transparency of data-driven models are essential in making decisions \citep{coorey2022health,ferrari2024digital}. Nevertheless, it remains challenging to discover the underlying physical system and the governing mechanism from data. Taking the material modeling task for instance, given that only the deformation field is observable, the goal of discovering the underlying material parameter field and mechanism presents an ill-posed unsupervised learning task. That means, even if an ML model can serve as a good surrogate to predict the corresponding loading field from a given deformation field, its inference of the material parameters can still drastically deteriorate.

\begin{wrapfigure}{r}{0.5\textwidth}
\vspace{-\intextsep}
\includegraphics[width=1\linewidth]
{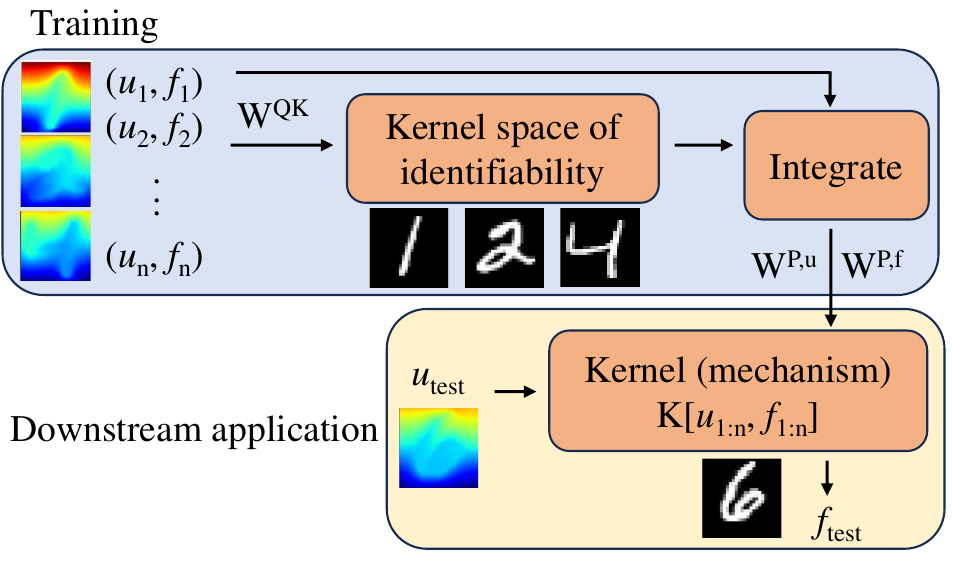}
 \caption{{\small Illustration of NAO's architecture.}}
 \label{fig:nao_architecture}
\end{wrapfigure}

To discover an interpretable mechanism for physical systems, a major challenge is to infer the governing laws of these systems that are often high- or infinite-dimensional, from data that are comprised of discrete measurements of continuous functions.
Therefore, a data-driven surrogate model needs to learn not only the mapping between input and output function pairs, but also the mapping from given function pairs to the hidden state. From the PDE-based modeling standpoint, learning a surrogate model corresponds to a forward problem, whereas inferring the underlying mechanism is an inverse problem. The latter is generally an enduring ill-posed problem, especially when the measurements are scarce.  
Unfortunately, such an ill-posedness issue may become even more severe in neural network models, due to the inherent bias of neural network approximations \citep{xu2019frequency}. To tackle this challenge, many deep learning methods have recently been proposed as inverse PDE solvers \citep{fan2023solving,molinaro2023neural,jiang2022reinforced,chen2023let}. 
The central idea is to incorporate prior information into the learning scheme, in the form of governing PDEs \citep{yang2021b,li2021physics}, regularizers \citep{dittmer2020regularization,obmann2020deep,ding2022coupling,chen2023let}, or additional operator structures \citep{uhlmann2009electrical,lai2019inverse,yilmaz2001seismic}. However, such prior information is often either unavailable or problem-specific in complex systems. As a result, these methods can only solve the inverse problem for a particular system, and one has to start from scratch when the system varies (e.g., when the material of the specimen undergoes degradation in a material modeling task).

In this work, we propose \textbf{Nonlocal Attention Operator (NAO)}, a novel attention-based neural operator architecture to simultaneously solve both forward and inverse modeling problems. Neural operators (NOs) \citep{li2020neural,li2020fourier} learn mappings between infinite-dimensional function spaces in the form of integral operators, hence they provide promising tools for the discovery of continuum physical laws by manifesting the mapping between spatial and/or spatiotemporal data; see \cite{you2022nonlocal,liuharnessing,liu2024domain,liu2023ino,Ong2022,cao2021choose,lu2019deeponet,lu2021learning,goswami2022physics,gupta2021multiwavelet} and references therein. However, most NOs focus on providing an efficient surrogate for the underlying physical system as a forward solver. They are often employed as black-box universal approximators but lack interpretability of the underlying physical laws. In contrast, the key innovation of NAO is that it introduces a kernel map based on the attention mechanism for simultaneous learning of the operator and the kernel map. As such, the kernel map automatically infers the context of the underlying physical system in an unsupervised manner. Intuitively, the attention mechanism extracts hidden knowledge from multiple systems by providing a function space of identifiability for the kernels, which acts as an automatic data-driven regularizer and endows the learned model's generalizability to new and unseen system states.

In this context, NAO learns a kernel map using the attention mechanism and simultaneously solves both the forward and inverse problems. The kernel map, whose parameters extract the global information about the kernel from multiple systems, efficiently infers resolution-invariant kernels from new datasets. As a consequence, NAO can achieve interpretability of the nonlocal operator and enable the discovery of hidden physical laws. \textbf{Our key contributions} include: 
\begin{itemize}
\item We bridge the divide between inverse PDE modeling and physics discovery tasks, and present a method to simultaneously perform physics modeling (forward PDE) and mechanism discovery (inverse PDE).
\item We propose a novel neural operator architecture NAO, based on the principle of contextual discovery from input/output function pairs through a kernel map constructed from multiple physical systems. As such, NAO is generalizable to new and unseen physical systems, and offers meaningful physical interpretation through the discovered kernel.
\item We provide theoretical analysis to show that the attention mechanism in NAO acts to provide the space of identifiability for the kernels from the training data, which reveals its ability to resolve ill-posed inverse PDE problems.
\item We conduct experiments on zero-shot learning to new and unseen physical systems, demonstrating the generalizability of NAO in both forward and inverse PDE problems.
\end{itemize}


\section{Background and related work}


Our work resides at the intersection of operator learning, attention-based models, and forward and inverse problems of PDEs. The ultimate goal is to model multiple physical systems from data while simultaneously discovering the hidden mechanism.

\textbf{Neural operator for hidden physics.}
Learning complex physical systems directly from data is ubiquitous in scientific and engineering applications   \citep{ghaboussi1998autoprogressive,liu2024deep,ghaboussi1991knowledge,carleo2019machine,karniadakis2021physics,zhang2018deep,cai2022physics,pfau2020ab,he2021manifold,besnard2006finite}. In many applications, the underlying governing laws are unknown, hidden in data to be revealed by physical models. Ideally, these models should be \emph{interpretable} for domain experts, who can then use these models to make further predictions and expand the understanding of the target physical system. 
Also, these models should be \emph{resolution-invariant}. Neural operators are designed to learn mappings between infinite-dimensional function spaces \citep{li2020neural,li2020multipole,li2020fourier,you2022nonlocal,Ong2022,cao2021choose,lu2019deeponet,lu2021learning,goswami2022physics, gupta2021multiwavelet}. As a result, NOs provide a promising tool for the discovery of continuum physical laws by manifesting the mapping between spatial and/or spatio-temporal data. 

\textbf{Forward and inverse PDE problems.} Most current NOs focus on providing an efficient surrogate for the underlying physical system as a forward PDE solver. They are often employed as black-box universal approximators without interpretability of the underlying physical laws. Conversely, several deep learning methods have been proposed as inverse PDE solvers \citep{fan2023solving,molinaro2023neural,jiang2022reinforced,chen2023let}, aiming to reconstruct the parameters in the PDE from solution data. Compared to the forward problem, the inverse problem is typically more challenging due to its ill-posed nature. To tackle the ill-posedness, many  NOs incorporate prior information, in the form of governing PDEs \citep{yang2021b,li2021physics}, regularizers \citep{dittmer2020regularization,obmann2020deep,ding2022coupling,chen2023let}, or additional operator structures \citep{uhlmann2009electrical,lai2019inverse,yilmaz2001seismic}. To our knowledge, our NO architecture is the first that solves both the forward (physics prediction) and inverse (physics discovery) problems simultaneously.

\textbf{Attention mechanism.} Since 2017, the attention mechanism has become the backbone of state-of-the-art deep learning models on many core AI tasks like NLP and CV. By calculating the similarity among tokens, the attention mechanism captures long-range dependencies between tokens \citep{vaswani2017attention}. Then, the tokens are spatially mixed to obtain the layer output. Based on the choice of mixers, attention-based models can be divided into three main categories: discrete graph-based attentions \citep{child2019generating,ho2019axial,wang2020linformer,katharopoulos2020transformers}, MLP-based attentions \citep{tolstikhin2021mlp,touvron2022resmlp,liu2021pay}, and convolution-based attentions \citep{lee2021fnet,rao2021global,guibas2021adaptive,nekoozadeh2023multiscale}. While most attention models focus on discrete mixers, it is proposed in \cite{guibas2021adaptive,nekoozadeh2023multiscale,tsai2019transformer,cao2021choose,wei2023super} to frame token mixing as a kernel integration, with the goal of obtaining predictions independent of the input resolution. 

Along the line of PDE-solving tasks, various attention mechanisms have been used to enlarge model capacity.  To improve the accuracy of forward PDE solvers, \cite{cao2021choose} removes the softmax normalization in the attention mechanism and employs linear attention as a learnable kernel in NOs. Further developments include the Galerkin-type linear attention in an encoder-decoder architecture in OFormer \citep{li2022transformer}, a hierarchical transformer for learning multiscale problems \citep{liu2022ht}, and a heterogeneous normalized attention with a geometric gating mechanism \citep{hao2023gnot} to handle multiple input features. In particular, going beyond solving a single PDE, the foundation model feature of attention mechanisms has been applied towards solving multiple types of PDEs within a specified context in \cite{yang2024pde,ye2024pdeformer,sun2024towards,zhang2024modno}. However, to our best knowledge, none of the existing work discovers hidden physics from data, nor do they discuss the connections between the attention mechanism and the inverse PDE problem.



\section{Nonlocal Attention Operator}


Consider multiple physical systems that are described by a class of operators mapping from input functions $u\in\mathbb{X}$ to output functions $f\in\mathbb{Y}$. Our goal is to learn the common physical law, in the form of operators $\mathcal{L}_{K}: \mathbb{X}\to \mathbb{Y}$ with system-dependent kernels $K$: 
 \begin{equation}\label{coarsegrained}
\mathcal{L}_K[u] + \epsilon = f. 
\end{equation}
Here $\mathbb{X}$ and $\mathbb{Y}$ are Banach spaces, 
$\epsilon $ denotes an additive noise describing the discrepancy between the ground-truth operator and the optimal surrogate operator, and $K$ is a kernel function representing the nonlocal spatial interaction. As such, the kernel provides the knowledge of its corresponding system, while \eqref{coarsegrained} offers a zero-shot prediction model for new and unseen systems.

To formulate the learning, we consider $n_{train}$ training datasets from different systems, with each dataset containing $d_u$ function pairs $(u,f)$: 
\begin{equation}\label{eq:data_training}
\mathcal{D}_{\rm tr} = \{\{({u}^\eta_i(x),f^\eta_i(x))\}_{i=1}^{d_u}\}_{\eta=1}^{n_{train}}.  
\end{equation}
In practice, the data of the input and output functions are on a spatial mesh $\{x_k\}_{k=1}^{n_x}\subset \Omega \subset\real^{d_x}$. The $n_{train}$ models with kernels $\{K^\eta\}_{\eta=1}^{n_{train}}$ correspond to different material micro-structures or different parametric settings. As a demonstration, we consider models for heterogeneous materials with operators in the form 
 \begin{equation}\label{eq:mixer}
 \mcL_K[u](x) = \int_\Omega K(x, y) g[u](y)dy, \, x\in \Omega,
 \end{equation}
 where $g[u](y)$ is a functional of $u$ determined by the operator; for example $g[u](y)= u(y)$ in Section \ref{sec:hetero}. Our approach extends naturally to other forms of operators, such as those with radial interaction kernels in Section \ref{sec:nonlocal_kernel} and heterogeneous interaction in the form of $ \mcL_K[u](x) = \int_\Omega K(x, y) g[u](x,y)dy$. Additionally, for simplicity, we consider scalar-valued functions $u$ and $f$ and note that the extension to vector-valued functions is trivial. 

\textbf{Remark: }Such an operator learning problem arises in many applications in forward and inverse PDE-solving problems. The inference of the kernel $K$ is an inverse problem, and the learning of the nonlocal operator is a forward problem. When considering a single physical system and taking $K$ in \eqref{eq:mixer} as an input-independent kernel, classical NOs can be obtained for forward PDE-solving tasks \citep{li2020fourier,guibas2021adaptive} and governing law learning tasks \citep{you2020data,jafarzadeh2023peridynamic}. Different from existing work, we consider the operator learning across multiple systems.


\subsection{Kernel map with attention mechanism}


The key ingredient in NAO is a kernel map constructed using the attention mechanism. It maps from data pairs to an estimation of the underlying kernel. The kernel map 
\begin{equation}\label{eq:K_attention}
\{(u_i,f_i)\}_{i=1}^{d_u}\, \to\, K[\mathbf{u}_{1:d},\mathbf{f}_{1:d};\theta]
\end{equation}
 has parameters $\theta$ estimated from the training dataset \eqref{eq:data_training}. As such, it maps from the token $(\mathbf{u}_{1:d},\mathbf{f}_{1:d})$ of the dataset $\{(u_i,f_i)\}_{i=1}^{d_u}$ to a kernel estimator, acting as an inverse PDE solver. 

A major innovation of this kernel map is its dependence on both $u$ and $f$ through their tokens. Thus, our approach distinguishes itself from the forward problem-solving NOs in the related work section, where the attention depends only on $u$. 


We first transfer the data $\{(u_i,f_i)\}_{i=1}^{d_u}$ to tokens $(\mathbf{u}_{1:d},\mathbf{f}_{1:d})$ according to the operator in \eqref{eq:mixer} by 
 \begin{equation}\label{eq:u_to_token}
\begin{aligned} 
\mathbf{u}_{1:d} &= (\mathbf{u}_{1}, \ldots,\mathbf{u}_{d})  = \big( g[u_j](y_{k})\big)_{1\leq j\leq d, 1\leq k\leq N}  \in \real^{N\times d}, \quad \\
\mathbf{f}_{1:d} & = (f_j(x_k))_{1\leq j\leq d, 1\leq k\leq N}\in \real^{N\times d},  
\end{aligned}
\end{equation}
where $d=d_u$ and $N=n_x$, assuming that $g[u]$ has a spatial mesh $\{y_k=x_k\}_{k=1}^N$.

Then, our discrete $(L+1)$-layer attention model for the inverse PDE problem writes:
{\small\begin{equation}\label{eq:K-attn_Layer}
\begin{split}
\Xb_{\text{in}}&=\Xb^{(0)}= (\Ub^{(0)},\Fb^{(0)}):= (\mathbf{u}_{1:d};\mathbf{f}_{1:d})  \in\real^{2N\times d}, 
\\
\Xb^{(l)}&=\text{Attn}[\Xb^{(l-1)};\theta_l]\Xb^{(l-1)}+\Xb^{(l-1)}=:(\Ub^{(l)},\Fb^{(l)})\in\real^{2N\times d}, \quad 1\leq l< L,   \\
\Xb_{\text{out}}&=\Xb^{L}=K[\mathbf{u}_{1:d},\mathbf{f}_{1:d};\theta]\mathbf{u}_{1:d}\approx \mathbf{f}_{1:d}\in\real^{N\times d}. 
\end{split}
\end{equation}}
Here, $\theta_l= (\Wb_{l}^Q\in\real^{d\times d_k}, \Wb_{l}^K\in\real^{d\times d_k})$ and the attention function is  
{\small \[
 \text{Attn}[\Xb;\theta_l]=\sigma\left(\frac{1}{\sqrt{d_k}} \Xb \Wb_{l}^Q {\Wb_{l}^K}^\top \Xb^\top\right) \in \real^{2N\times 2N}. \]
 }
The kernel map is defined as 
{\small
\begin{equation}\label{eq:kernel_map}
\begin{aligned}
K[\mathbf{u}_{1:d},\mathbf{f}_{1:d};\theta]& =W^{P,u} \sigma\left(\frac{1}{\sqrt{d_k}}(\Ub^{(L-1)})^\top \Wb_L^Q (\Wb_L^K)^\top \Ub^{(L-1)}  \right)  \\
&+ W^{P,f}\sigma\left(\frac{1}{\sqrt{d_k}} (\Fb^{(L-1)})^\top \Wb_L^Q  (\Wb_L^K)^\top \Ub^{(L-1)} \right),
\end{aligned}
\end{equation}
}
where $\theta:=\{W^{P,u}\in\real^{N\times N}, W^{P,f}\in\real^{N\times N}, \{\Wb_{l}^Q, \Wb_{l}^K\}_{l=1}^{L}\}$ are learnable parameters. Here, we note that $d_k$ only controls the rank bound when lifting each point-wise feature via $\Wb_{l}^Q(\Wb_{l}^K)^\top$. In the following, we denote $\Wb^{QK}_l:=\frac{1}{\sqrt{d_k}}\Wb_{l}^Q(\Wb_{l}^K)^\top$, which characterizes the (trainable) interaction of $d$ input function pair instances.


\subsection{Nonlocal Attention Operator in continuum limit}


As suggested by \cite{cao2021choose}, we take the activation function $\sigma$ as a linear operator. Then, noting that the matrix multiplication in \eqref{eq:K-attn_Layer} is a Riemann sum approximation of an integral (with a full derivation in Appendix~\ref{sec:Riemann}), we propose the nonlocal attention operator as the continuum limit of \eqref{eq:K-attn_Layer}:
{\small\begin{align}
\nonumber&g_j^{(0)}(x):=g[u_j](x),\; f_j^{(0)}(x):=f_j(x),\\
\nonumber&\left(\begin{array}{c}
g_j^{(l)}(x)\\
f_j^{(l)}(x)\\
\end{array}\right)=\int_\omg K^{(l)}(x,y)
\left(\begin{array}{c}
g_j^{(l-1)}(y)\\
f_j^{(l-1)}(y)\\
\end{array}\right)dy+
\left(\begin{array}{c}
g_j^{(l-1)}(x)\\
f_j^{(l-1)}(x)\\
\end{array}\right),\quad 1\leq l<L,\\
&\mcL_{K[\mathbf{u}_{1:d},\mathbf{f}_{1:d};\theta]}[{u}](x)= \int_\Omega K[\mathbf{u}_{1:d},\mathbf{f}_{1:d};\theta](x,y) g[u](y)dy,\label{eqn:metamodel_full}
\end{align}}
in which the integration is approximated by the Riemann sum in our implementation. Here,
{\small\begin{align*}
\nonumber&K^{(l)}(x,y):=\left[\begin{array}{cc}
\sum_{\omega,\nu=1}^{d}\left( g^{(l-1)}_\omega(x) \Wb_{l}^{QK}[\omega,\nu] g^{(l-1)}_\nu(y)\right)&\sum_{\omega,\nu=1}^{d}\left( g^{(l-1)}_\omega(x) \Wb_{l}^{QK}[\omega,\nu] f^{(l-1)}_\nu(y)\right)\\
\sum_{\omega,\nu=1}^{d}\left( f^{(l-1)}_\omega(x) \Wb_{l}^{QK}[\omega,\nu] g^{(l-1)}_\nu(y)\right)&\sum_{\omega,\nu=1}^{d}\left( f^{(l-1)}_\omega(x) \Wb_{l}^{QK}[\omega,\nu] f^{(l-1)}_\nu(y)\right)\\
\end{array}\right]\\
&K[\mathbf{u}_{1:d},\mathbf{f}_{1:d};\theta](x,y):=\sum_{\omega,\nu=1}^{d}\int_{\omg} W^{P,u}(x,z) \left(g^{(L-1)}_{\omega}(z) W_L^{QK}[\omega,\nu] g^{(L-1)}_{\nu}(y)  \right)dz\\
&\qquad\qquad\qquad\qquad+\sum_{\omega,\nu=1}^{d}\int_{\omg} W^{P,f}(x,z) \left(f^{(L-1)}_{\omega}(z) W_L^{QK}[\omega,\nu] g^{(L-1)}_{\nu}(y)  \right)dz.
\end{align*}}
We learn the parameters $\theta$ 
by minimizing the following mean squared error loss function:
{\small\begin{equation}\label{eq:lossFn}
\mcE(\theta):=\frac{1}{n_{train}} \sum_{\eta=1}^{n_{train}} \sum_{i=1}^d \int_{\omg}\verti{\mcL_{K[\mathbf{u}^\eta_{1:d},\mathbf{f}^\eta_{1:d};\theta]} [u_i^\eta](x)-f_i^\eta(x)}^2 dx.
\end{equation}}
The performance of the model is evaluated on test tasks with new and unseen kernels ${\mcD}_{test}=\{({u}^{test}_i(x),f^{test}_i(x))\}_{i=1}^d$ based on the following two criteria:
{\small\begin{align}
\text{Operator (forward PDE) Error: }&E_{f}:=\frac{1}{d}\sum_{i=1}^d \frac{\vertii{\mcL_{K[\mathbf{u}^{test}_{1:d},\mathbf{f}^{test}_{1:d} ;\theta]} [u_i^{test}](x)-f_i^{test}(x)}_{L^2(\omg)}}{\vertii{f_i^{test}(x))}_{L^2(\omg)}},\label{eqn:operator_err}\\
\text{Kernel (inverse PDE) Error: }&E_{K}:=\verti{ K[\mathbf{u}^{test}_{1:d},\mathbf{f}^{test}_{1:d};\theta]-K_{test}}_{L^2(\rho)}/\verti{K_{test}}_{L^2(\rho)}.\label{eqn:kernel_err}
\end{align}}
Here, $L^2(\rho)$ is the empirical measure in the kernel space as defined in \cite{lu2023nonparametric}. Note that $\theta$ are trained using multiple datasets. Intuitively speaking, the attention mechanism helps encode the prior information from other tasks for learning the nonlocal kernel, and leads to estimators significantly better than those using a single dataset. To formally understand this mechanism, we analyze a shallow 2-layer NAO in the next section.

\section{Understanding the attention mechanism}\label{sec:analysis}


To facilitate further understanding of the attention mechanism, we analyze the limit of the two-layer attention-parameterized kernel in \eqref{eq:kernel_map} and the range of the kernel map, which falls in the space of identifiability for the kernels from the training data. We also connect the kernel map with the regularized estimators. 
For simplicity, we consider operators of the form 
 \begin{equation}\label{eq:radial_K}
 \mcL_K[u](x) = \int_0^\delta K(r) g[u](r,x)dr, \, x\in \Omega,
 \end{equation} 
which is the radial nonlocal kernel in Sec.\ref{sec:nonlocal_kernel}. 

 \subsection{Limit of the two-layer attention-parameterized kernel}


 We show that as the number $N$ and the spatial mesh $n_x$ approach infinity, the limit of the two-layer attention-parameterized kernel is a double integral. Its proof is in Appendix \ref{sec:appd-proofs}. 
 
For simplicity, we assume that the dataset in \eqref{eq:data_training} has $d_u=1$ and $n_{train}=1$  with a uniform mesh $\{x_j\}_{j=1}^{n_x}$. We define the tokens by 
 \begin{equation}\label{eq:u_to_token2}
\begin{aligned} 
\mathbf{u}_{1:d} =  (\mathbf{u}_{1}, \ldots,\mathbf{u}_{d}) 
=\left( g[u](r_k,x_{j} ) \right)_{1\leq j\leq d, 1\leq k\leq N}  \in \real^{N \times d},\;  \mathbf{f}_{1:d}  = (f(x_{j}))_{1\leq j\leq d} \in \real^{1\times d}, 
\end{aligned}
\end{equation}
where $d=n_x$ and $\{r_k\}_{k=1}^N$ is the spatial mesh for $K$'s independent variable $r\in [0,\delta]$.

 \begin{lemma}\label{lemma:attn-limit}
Consider the two-layer attention model in \eqref{eq:K-attn_Layer}--\eqref{eq:kernel_map} with bounded parameters. For each $d$ and $N$, let $\{x_{j}\}_{j=1}^d$ and $\{r_k\}_{k=1}^N$ be a uniform meshes of the compact sets $\Omega$ and $[0,\delta]$, and let $\{A_j\}_{j=1}^d$ be the resulting uniform partition of $\Omega$. Assume that $g[u]$ in \eqref{eq:radial_K} is continuous on $[0,\delta]\times \Omega$. 
Then,   
{\small  \begin{equation}\label{eq:att_limit}
\begin{aligned}
 & \lim_{N\to\infty}  \lim_{d\to\infty} \sum_{k=1}^N K[\mathbf{u}_{1:d},\mathbf{f}_{1:d};\theta](r_k)\mathbf{1}_{[r_{k-1},r_{k})}(r) (r_{k}-r_{k-1}) 
\\
 = &  K[u,f](r):=\int_{0}^\delta W^{P,u}(|r'|)\sigma\left(\int \int \left[g[u](r',x) W^{QK}(x,y)  g[u](r,y)  dxdy\right]\right) dr' \\
&+W^{P,f}\sigma\left(\int\int \left[f(x)W^{QK}(x,y)  g[u](r,y)  \right]dxdy\right), 
\end{aligned}
\end{equation}}
where $W^{QK}(x,y) =\lim_{d\to \infty} \sum_{j,j'=1}^d W^{QK}[j,j']\mathbf{1}_{A_j\times A_{j'}}(x,y) $ 
is the scaled $L^2(\Omega\times \Omega)$ limit of the parameter matrix $W^{QK}[j',j] =\sum_{l=1}^{d_k} W^Q[j,l]\cdot W^K[j',l]$ and $W^{P,u}(r) =\lim_{N\to \infty} \sum_{k=1}^N W^{P,u}[k]\mathbf{1}_{[r_{k-1},r_{k})}(r) $. 
 \end{lemma}


\subsection{Space of identifiability for the kernels}


For a given training dataset, we show that the function space in which the kernels can be identified is the closure of a data-adaptive reproducing kernel Hilbert space (RKHS). This space contains the range of the kernel map and hence provides the ground for analyzing the inverse problem. 

\begin{lemma}[Space of Identifiability]\label{lemma:ID}
	Assume that the training data pairs in \eqref{eq:data_training} are sampled from continuous functions $\{u_i^\eta\}_{i,\eta=1}^{d_u,n_{train}}$ with a compact support. Then, the function space the loss function in \eqref{eq:lossFn} has a unique minimizer $K(s) = K[\mathbf{u}^\eta_{1:d},\mathbf{f}^\eta_{1:d};\theta](s)$ is the closure of a data-adaptive RKHS $H_{G}$ with a reproducing kernel $\bar{G}$ determined by the training data: 
	$$\bar{G}(r,s)=[\rho'(r)\rho'(s)]^{-1}G(r,s),$$
 where $\rho'$ is the density of the empirical measure $\rho$ defined by   
 \begin{align}
 \rho'(r) :=\dfrac{1}{Z}\sum_{\eta=1}^{n_{train}} \sum_{i=1}^{d_u}\int_{\omg} | g[u_i^\eta](r,x) | dx, \label{eq:rho}
 \end{align}
 and the function $G$ is defined by 
 $ G(r,s)
  :=  \dfrac{1}{n_{train}d} \sum_{\eta=1}^{n_{train}} \sum_{i=1}^{d_u} \int_{\omg} g[u_i^\eta](r,x) g[u_i^\eta](s,x) dx.    
$ 
\end{lemma}

The above space is data-adaptive since the integral kernel $\bar{G}$ depends on data. It characterizes the information in the training data for estimating the nonlocal kernel $K(s) = K[\mathbf{u}^\eta_{1:d},\mathbf{f}^\eta_{1:d};\theta](s)$. In general, the more data, the larger the space is. On the other hand, note that the loss function's minimizer with respect to $K(s)$ is not the kernel map. The minimizer is a fixed estimator for the training dataset and does not provide any information for estimating the kernel from another dataset.

\textbf{Comparison with regularized estimators.} 
The kernel map solves the ill-posed inverse problem using prior information from the training dataset of multiple systems, which is not used in classical inverse problem solvers. To illustrate this mechanism, consider the extreme case of estimating the kernel in the nonlocal operator from a dataset consisting of only a single function pair $(u,f)$. This inverse problem is severely ill-posed because of the small dataset and the need for deconvolution to estimate the kernel. Thus, regularization is necessary, where two main challenges present: (i) the selection of a proper regularization with limited prior information, and (ii) the prohibitive computational cost of solving the resulting large linear systems many times. 

In contrast, our kernel map $K[\mathbf{u}_{1:d},\mathbf{f}_{1:d};\theta](s)$, with the parameter $\theta$ estimated from the training datasets, acts on the token $(\mathbf{u}_{1:d},\mathbf{f}_{1:d})$ of $(u,f)$ to provide an estimator. It passes the prior information about the kernel from the training dataset to the estimation for new datasets. Importantly, it captures the nonlinear dependence of the estimator on the data $(u,f)$.  Computationally, it can be applied directly to multiple new datasets without solving the linear systems.  
In Section \ref{sec:appd-reguEst}, we further show that a regularized estimator depends nonlinearly on the data pair $(u,f)$. In particular, similar to Lemma \ref{lemma:attn-limit}, there is an RKHS determined by the data pair $(u,f)$. The regularized estimator suggests that the kernel map can involve a component quadratic in the feature $g[u]$, similar to the limit form of the attention model in Lemma \ref{lemma:attn-limit}. 


\section{Experiments} \label{sec:experiments}


We assess the performance of NAO on a wide range of physics modeling and discovery datasets. Our evaluation focuses on several key aspects:
1) we demonstrate the merits of the continuous and linear attention mechanism, compare the performance with the baseline discrete attention model (denoted as Discrete-NAO), the softmax attention mechanism (denoted as Softmax-NAO), with input on $u$ only (denoted as NAO-u), the convolution-based attention mechanism (denoted as AFNO \citep{guibas2021adaptive}), and an MLP-based encoder architecture that maps the datum $[\ub_{1:d},\fb_{1:d}]$ directly to a latent kernel (denoted as Autoencoder); 2) we measure the generalizability, in particular, the zero-shot prediction performance in modeling a new physical system with unseen governing equations, and across different resolutions; 3) we evaluate the data efficiency-accuracy trade-off in ill-posed inverse PDE learning tasks, as well as the interpretability of the learned kernels. In all experiments, the optimization is performed with the Adam optimizer. To conduct fair comparison for each method, we tune the hyperparameters, including the learning rates, the decay rates, and the regularization parameters, to minimize the training loss. In all examples, we use 3-layer models, and parameterize the kernel network $W^{P,u}$ and $W^{P,f}$ with a 3-layer MLP with hidden dimensions $(32,64)$ and LeakyReLU activation. Experiments are conducted on a single NVIDIA GeForce RTX 3090 GPU with 24 GB memory. Additional details on data generation and training strategies are provided in Appendix \ref{sec:exp_more}.


\subsection{Radial kernel learning}\label{sec:nonlocal_kernel}


{\small\begin{table*}[h!]
\caption{{Test errors and the number of trainable parameters for the radial kernel problem, where bold numbers highlight the best methods. The small operator errors and large kernel errors of discrete-NAO highlight the ill-posedness of the inverse problem. NAO overcomes the ill-posedness and yields resolution-invariant estimators.     
}}
\label{tab:kernel_results}
\begin{center}
{\small    \centering
\begin{tabular}{cc|r|cc|cc}
\hline
Case & model & \#param &\multicolumn{2}{c}{Operator test error} &\multicolumn{2}{c}{Kernel test error}\\
\cline{4-7}
&&&ID & OOD1 &ID & OOD1 \\
\hline\hline
& Discrete-NAO &  16526 & {\bf 1.33\%} & 25.81\% & 29.02\% &28.80\%\\
&Softmax-NAO &  18843 &
13.45\%& 66.06\% & 67.55\% & 85.80\%\\
$d$ = 302, $d_k$=10&AFNO & 19605 & 22.62\% & 68.76\% & - & -
\\
&NAO &  18843 & 1.48\%& {\bf 8.10\%} & {\bf 5.40\%}& {\bf 10.02\%}\\
&NAO-u &  18842 & 13.68\% & 66.68\% & 20.46\% & 74.03\%\\
& Autoencoder &  16424 & 12.97\%& { 1041.49\%} & { 22.56\%}& { 136.79\%}\\
\hline\hline
$d$ = 302, $d_k$=5&Discrete-NAO & 10465  & {\bf 1.63\%}& 15.80\%&33.21\%&30.39\%\\
&NAO & 12783 & 2.34\% & {\bf 9.23\%} &{\bf 6.87\%}&{\bf 14.62\%}\\
\hline
$d$ = 302, $d_k$=20&Discrete-NAO & 28645 & 1.35\%& 18.70\% & 35.49\% &30.81\%\\
&NAO & 30963 & {\bf 1.33\%} & {\bf 9.12\%} &{\bf 4.63\%}& {\bf 9.14\%}\\
\hline\hline
$d$ = 100, $d_k$=10&Discrete-NAO & 8446  & 1.73\% & 14.92\% &34.52\% & 35.20\%\\
&NAO & 10763 & {\bf 1.07\%} & {\bf 6.35\%} & {\bf 7.41\%} & {\bf 17.02\%}\\
\hline
$d$ = 50, $d_k$=10&Discrete-NAO & 6446  & 2.29\% & 10.31\%& 41.80\% & 45.30\%\\
&NAO & 8763 & {\bf 1.56\%} & {\bf 7.19\%}& {\bf 15.95\%} & {\bf 29.47\%}\\
\hline
$d$ = 30, $d_k$=10&Discrete-NAO & 5646  & 5.60\% & 11.31\% & 58.24\% & 64.23\%\\
&NAO & 7963 & {\bf 2.94\%} & {\bf 8.04\%} & {\bf 22.65\%} & {\bf 33.77\%}\\
\hline\hline
\end{tabular}}
\end{center}
\vskip -0.15in
\end{table*}}

In this example, we consider the learning of nonlocal diffusion operators, in the form:
{\small\begin{equation}\label{eq:nonlocal_op}
\mcL_{\gamma_\eta}[u](x) =  \int_{\Omega} \gamma_\eta(\vert y-x\vert) [u(y)-u(x)]dy= f(x), \forall x\in \Omega. 
\end{equation}} 
Unlike a (local) differential operator, this operator depends on the function $u$ nonlocally through the convolution of $u(y)-u(x)$, and the operator is characterized by a radial kernel $\gamma_\eta$. It finds broad physical applications in describing fracture mechanics \citep{silling2000reformulation}, anomalous diffusion behaviors \citep{bucur2016nonlocal}, and the homogenization of multiscale systems \citep{du2020multiscale}. 

In this context, our goal is to learn the operator $\mcL$ as well as to discover the hidden mechanism, namely the kernel $K[\ub_{1:d},\fb_{1:d};\theta](x,y)=\gamma_\eta(\verti{y-x})$. In the form of the operator in \eqref{eq:radial_K}, we have $K(r)= \gamma_{\eta}(r)$ and  
$g[u](r,x) = u(x+r) + u(x-r) - 2u(x) 
$
for $  r\in [0,\delta]$.

To generate the training data, we consider $7$ sine-type kernels
\begin{equation}\label{eqn:sine_kernel}
\gamma_{\eta}(\verti{y-x}):=\exp(-\eta(\verti{y-x}))\sin(6\verti{y-x})\mathbf{1}_{[0,11]}(\verti{y-x}),\;\eta=1,2,3,4,6,7,8.
\end{equation}
Here, $\eta$ denotes task index. We generate $4530$ data pairs $(g^\eta[u],f^\eta)$ with a fixed resolution $\Delta x=0.0125$ for each task, where the loading function $\mcL_{\gamma_\eta}[u^\eta]=f^\eta$ is computed by the adaptive Gauss-Kronrod quadrature method. Then, we form a training sample of each task by taking $d$ pairs from this task. When taking the token size $d=302$, each task contains $\frac{4530}{302}=15$ samples. We consider two test kernels: one following the same rule of \eqref{eqn:sine_kernel} with $\eta=5$ (denoted as the ``in-distribution (ID) test'' system), and the other following a different rule (denoted as the ``out-of-distribution (OOD) test1'' system):
\begin{equation}\label{eqn:sine_kernel_ood}
\gamma_{ood1}(\verti{y-x}):=\verti{y-x}(11-\verti{y-x})\exp(-5(\verti{y-x}))\sin(6\verti{y-x})\mathbf{1}_{[0,11]}(\verti{y-x}).
\end{equation}
Both the operator error \eqref{eqn:operator_err} and the kernel error \eqref{eqn:kernel_err} are provided in Table~\ref{tab:kernel_results}. While the former measures the error of the learned forward PDE solver (i.e., learning a physical model), the latter demonstrates the capability of serving as an inverse PDE solver (i.e., physics discovery).

\textbf{Ablation study. }We first perform an ablation study on NAO, by comparing its performance with its variants (Discrete-NAO, Softmax-NAO, and NAO-u), AFNO, and Autoencoder, with a fixed token dimension $d=302$, query-key feature size $d_k=10$, and data resolution $\Delta x=0.0125$. When comparing the operator errors, both Discrete-NAO and NAO serve as good surrogate models for the ID task with relative errors of $1.33\%$ and $1.48\%$, respectively, while the other three baselines show $>10\%$ errors. Therefore, we focus more on the comparison between Discrete-NAO and NAO. This gap becomes more pronounced in the OOD task: only NAO is able to provide a surrogate of $\mcL_{\gamma_{ood}}$ with $<10\%$ error, who outperforms its discrete mixer counterpart by $68.62\%$, indicating that NAO learns a more generalizable mechanism. This argument is further affirmed when comparing the kernel errors, where NAO substantially outperforms all baselines by at least $81.39\%$ in the ID test and $65.21\%$ in the OOD test. This study verifies our analysis in Section \ref{sec:analysis}: NAO learns the kernel map in the space of identifiability, and hence possesses advantages in solving the challenging ill-posed inverse problem. Additionally, we vary the query-key feature size from $d_k=10$ to $d_k=5$ and $d_k=20$. Note that $d_k$ determines the rank bound of $W^{QK}$, the matrix that characterizes the interaction between different data pairs. Discrete-NAO again performs well only in approximating the operator for the ID test, while NAO achieves consistent results in both tests and criteria, showing that it has successfully discovered the intrinsic low-dimension in the kernel space.
\begin{figure*}
    \centering
    \includegraphics[width=0.49\textwidth]{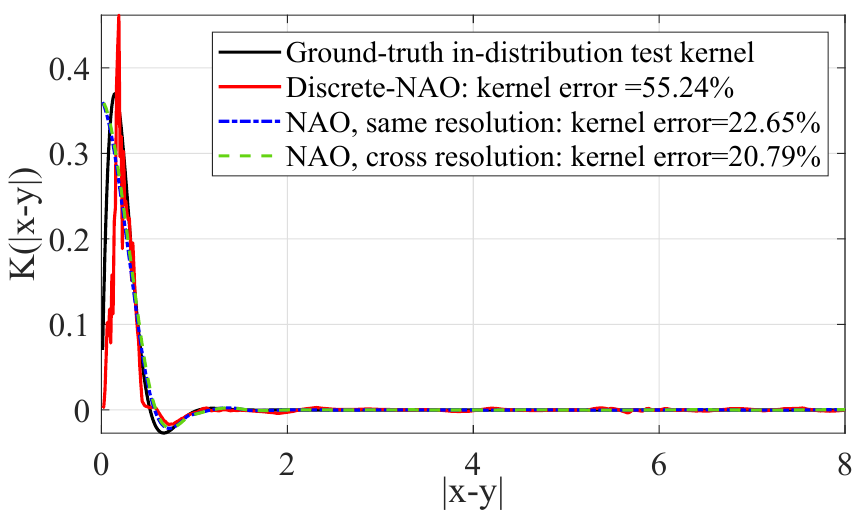}
    \hspace{0.008\textwidth}
    \includegraphics[width=0.49\textwidth]{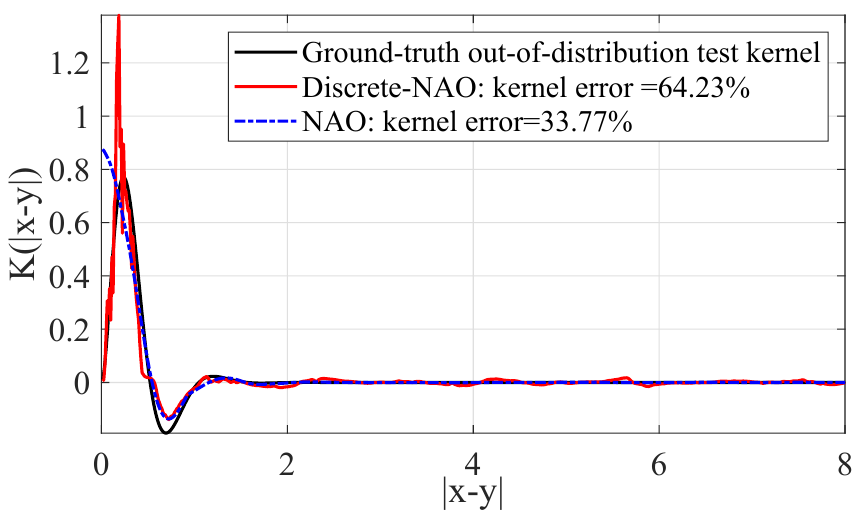}
    \caption{Results on radial kernel learning, when learning the test kernel from a small ($d=30$) number of data pairs: test on an ID task (left), and test on an OOD task (right).}
    \label{fig:kernel}
\end{figure*}

\textbf{Alleviating ill-posedness. } To further understand NAO's capability as an inverse PDE solver, we reduce the number of data pairs for each sample from $d=302$ to $d=30$, making it more ill-posed as an inverse PDE problem. NAO again outperforms its discrete mixer counterpart in all aspects. Interestingly, the errors in NAO increase almost monotonically, showing its robustness. For Discrete-NAO, the error also increases monotonically in the ID operator test, but there exists no consistent pattern in other test criteria. Figure \ref{fig:kernel} shows the learned test kernels in both the ID and OOD tasks. It shows that Discrete-NAO learns highly oscillatory kernels, while our continuous NAO only has a discrepancy near $|x-y|=0$. Note that when $|x-y|=0$, we have $u(y)-u(x)=0$ in the ground-truth operator \eqref{eq:nonlocal_op}, and hence the kernel value at this point does not change the operator value. That means, our data provides almost no information at this point. This again verifies our analysis: continuous NAO learns the kernel map structure from small data based on prior knowledge from other task datasets.

\textbf{Cross-resolution. } We test the NAO model trained with $\Delta x=0.0125$ on a dataset corresponding to $\Delta x=0.025$, and plot the results in Figure \ref{fig:kernel} Left. The predicted kernel is very similar to the one learned from the same resolution, and the error is also on-par ($22.65\%$ versus $20.79\%$).

\begin{table*}[t!]
\caption{{Training and test for the radial kernel problem with more diverse tasks with fixed $d=302$, where bold numbers highlight the best methods across different data settings. These results emphasize the balance between task diversity and the number of samples per task.}}
\vspace{-2mm}
\label{tab:kernel_results2}
\begin{center}
{\small    \centering
\begin{tabular}{cc|r|ccc|ccc}
\hline
$d_k$ & model & \#param &\multicolumn{3}{c}{Operator test error} &\multicolumn{3}{c}{Kernel test error}\\
\cline{4-9}
&&&ID & OOD1 & OOD2 &ID & OOD1 & OOD2 \\
\hline\hline
\multicolumn{9}{c}{{\bf sine only:} Train on sine dataset, 105 samples in total}\\
\hline
10& \small{Discrete-NAO} &  16526 & {1.33\%} & 25.81\% & 138.00\% & 29.02\% &28.80\% &97.14\%\\
&NAO &  18843 & 1.48\%& {8.10\%} & 221.04\% & {5.40\%}& {10.02\%} & 420.60\% \\
\hline
20& \small{Discrete-NAO} & 28645 & 1.35\%& 18.70\% & 99.50\% & 35.49\% &30.81\% &101.08\%\\
&NAO & 30963 & {1.33\%} & {9.12\%} & 211.78\% &{4.63\%}& {9.14\%} &329.66\%\\
\hline
40& \small{Discrete-NAO} & 52886 & {1.30\%}& 31.37\% & 49.83\% & 38.89\% & 30.02\% &129.84\% \\
&NAO & 55203 & {\bf 0.67\%} & {7.34\%} &234.43\% &{\bf 4.03\%}& {12.16\%}& 1062.3\%\\
\hline
& Autoencoder &  16424 & 12.97\%& { 1041.49\%} & 698.72\% & { 22.56\%}& { 136.79\%} & 304.37\% \\
\hline\hline
\multicolumn{9}{c}{{\bf sine+cosine+polyn:} Train on diverse (sine, cosine and polynomial) dataset, 315 samples in total}\\
\hline
10& \small{Discrete-NAO} &  16526 & {2.27\%} & {13.02\%} &11.50\% & 10.41\% & 30.79\% &77.80\% \\
&NAO &  18843 & 2.34\%& {14.37\%} & 10.05\% & {7.26\%}& {28.23\%} & 94.38\%\\
\hline
20& \small{Discrete-NAO} & 28645 & {1.60\%}& 6.03\% &19.73\% & 21.83\% &21.29\% & 18.97\%\\
&NAO & 30963 & {1.64\%} & {3.25\%} & {\bf 3.58\%} &{5.45\%}& {8.87\%} & 15.82\%\\
\hline
40& \small{Discrete-NAO} & 52886 & {1.45\%}& 5.49\% &18.26\% & 20.07\% &19.46\% &18.44\%\\
&NAO & 55203 & {1.54\%} & {\bf 3.09\%} &7.69\% &{5.04\%}& {\bf 6.92\%} & {\bf 10.48\%}\\
\hline
&Autoencoder &  56486 & 12.67\%& 341.96\% & 211.61\% & 27.06\% & 52.43\% & 128.08\% \\
\hline\hline
\multicolumn{9}{c}{{\bf Single task:} Train on a single sine dataset, 105 samples in total}\\
\hline
10& NAO &  18843 & 104.49\%& {104.37\%} &56.64\% & {100.31\%}& {100.00\%} &94.98\%\\
\hline
20&NAO & 30963 & {116.89\%} & {105.52\%} &85.40\%&{99.02\%}& {99.55\%} &98.37\%\\
\hline
40&NAO & 55203 & {111.33\%} & {104.12\%} & 76.78\% &{101.61\%}& {100.33\%}&95.90\%\\
\hline\hline
\multicolumn{9}{c}{{\bf Fewer samples:} Train on diverse (sine, cosine and polynomial) dataset, 105 samples in total}\\
\hline
10&NAO &  18843 & 4.23\% & {15.34\%} &11.13\% &{10.11\%}& {25.23\%} &97.63\%\\
\hline
20&NAO & 30963 & {4.15\%} & {10.15\%} &9.59\%&{8.84\%}& {23.08\%}&20.05\%\\
\hline
40&NAO & 55203 & {3.69\%} & {11.67\%} &6.08\% &{9.19\%}& {24.04\%} &25.58\%\\
\hline\hline
\end{tabular}}
\end{center}
\vskip -0.15in
\end{table*}

\textbf{More diverse tasks.} To further evaluate the generalization capability of NAO as a foundation model, we add another two types of kernels into the training dataset. The training dataset is now constructed based on 21 kernels of three groups, with $15$ samples on each kernel:
\begin{itemize}
\item sine-type kernels: $\gamma^{\sin}_\eta(r)=\exp(-\eta r)\sin(6r)\mathbf{1}_{[0,11]}(r)$, $\eta=1,2,3,4,6,7,8$.
\item cosine-type kernels: $\gamma^{\cos}_\eta(r)=\frac{10-r}{20}\cos(\eta r)(10-r)\mathbf{1}_{[0,10]}(r)$, $\eta=0,1,2,3,4,5,6$.
\item polynomial-type kernels: $\gamma^{\text poly}_\eta(r)=\exp(-0.1 r)p_\eta\left(\frac{r-10}{10}\right)\mathbf{1}_{[0,10]}(r)$, $\eta=1,2,3,4,5,6,7$, where $p_\eta$ is the degree-$\eta$ Legendre polynomial.
\end{itemize}

\begin{figure}[bt!]
\centering
\includegraphics[width=0.49\textwidth]{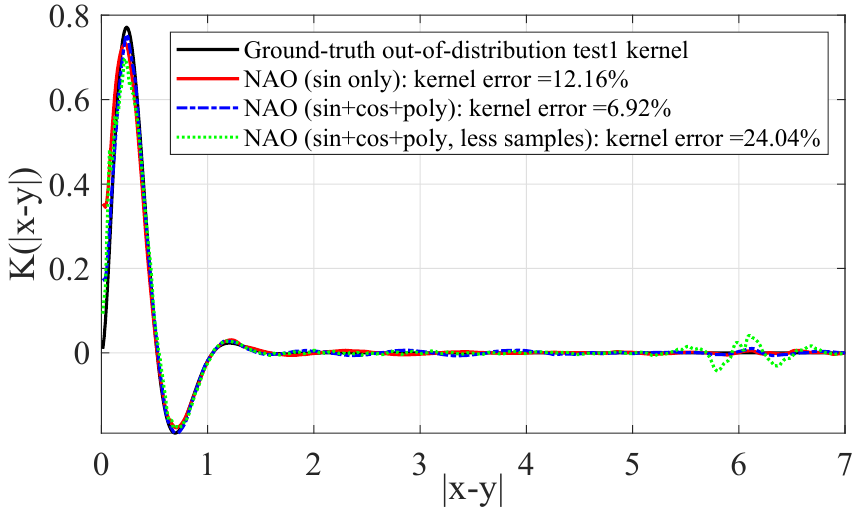}
\hspace{0.008\textwidth}
\includegraphics[width=0.49\textwidth]{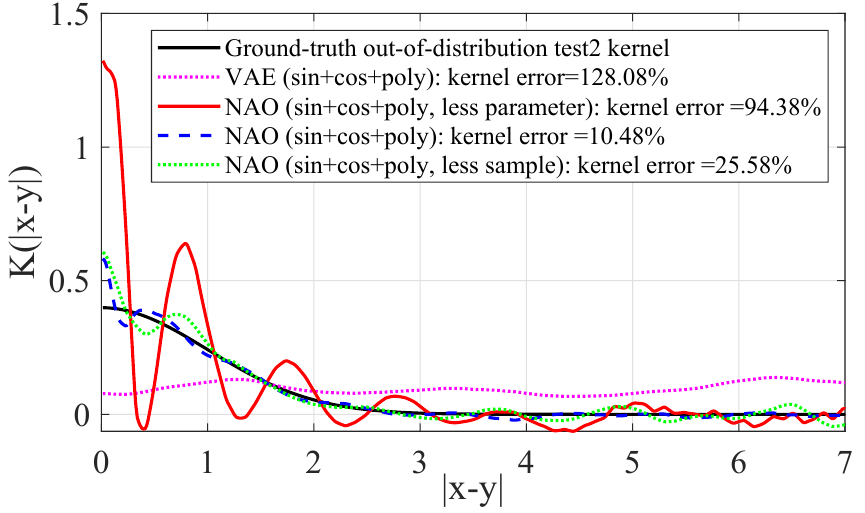}
\caption{OOD test results on radial kernel learning, with diverse training tasks and $d=302$. OOD1 (left): true kernel $\gamma(r)=r(11-r)\exp(-5r)\sin(6r)\mathbf{1}_{[0,11]}(r)$; OOD2 (right): true kernel $\gamma(r)=\frac{\exp(-0.5 r^2)}{\sqrt{2\pi}}$, a Gaussian kernel which is very different from all training tasks.}
\label{fig:kernel_results2}
\end{figure}

Based on this enriched dataset, besides the original ``sine only'' setting, three additional settings are considered to compare the generalizability across different settings. The results are reported in Table~\ref{tab:kernel_results2} and Figure~\ref{fig:kernel_results2}. In addition to the original setting corresponding to all ``sine'' kernels, in part II of Table~\ref{tab:kernel_results2} (denoted as ``sine+cosine+polyn''), we consider a ``diverse task'' setting, where all $315$ samples are employed in training. In part III, we consider a ``single task'' setting, where only the first sine-type kernel, $\gamma(r)=\exp(-r)\sin(6r)\mathbf{1}_{[0,11]}(r)$, is considered as the training task, with $105$ samples on this task. Lastly, in part IV we demonstrate a ``fewer samples'' setting, where the training dataset still consists of all $21$ tasks but with only $5$ samples on each task. For testing, besides the ID and OOD tasks in the ablation study, we add an additional OOD task with a Gaussian-type kernel $\gamma_{ood2}(r)=\frac{\exp(-0.5r^2)}{\sqrt{2\pi}}$, which is substantially different from all training tasks.

As shown in Table~\ref{tab:kernel_results2}, considering diverse datasets helps in both OOD tests (the kernel error is reduced from $9.14\%$ to $6.92\%$ in OOD test 1, and from $329.66\%$ to $10.48\%$ in OOD test 2), but not for the ID test (the kernel error is slightly increased from $4.03\%$ to $5.04\%$). We also note that since the ``sine only'' setting only sees systems with the same kernel frequency in training, it does not generalize to OOD test 2, where it becomes necessary to include more diverse training tasks. As the training tasks become more diverse, the intrinsic dimension of kernel space increases, requiring a larger rank size $d_k$ of the weight matrices. When comparing the  ``fewer samples'' setting with the ``sine only'' setting,  the former exhibits better task diversity but fewer samples per task. One can see that the performance deteriorates on the ID test but improves on OOD test 2. This observation emphasizes the balance between the diversity of tasks and the number of training samples per task.



\subsection{Solution operator learning}

We consider the modeling of 2D sub-surface flows through a porous medium with a heterogeneous permeability field. Following the settings in \cite{li2020neural}, the high-fidelity synthetic simulation data for this example are described by the Darcy flow. Here, the physical domain is $\omg=[0,1]^2$, $b(\xb)$ is the permeability field, and the Darcy's equation has the form:
\begin{align}
\label{eqn:darcy}-\nabla\cdot(b(\xb)\nabla p(\xb))=g(\xb),\;\quad\xb\in \omg;\qquad p(\xb)=0,\quad\xb\in\partial \omg.
\end{align}
In this context, we aim to learn the solution operator of Darcy's equation and compute the pressure field $p(\xb)$. We consider two study scenarios. 1) $g\rightarrow p$: each task has a fixed microstructure $b(\xb)$, and our goal is to learn the (linear) solution operator mapping from each loading field $g$ to the corresponding solution field $p$. In this case, the kernel $K$ acts as the Green's function of \eqref{eqn:darcy}, and can be approximated by the inverse of the stiffness matrix. 2) $b\rightarrow p$: each task has a fixed loading field $g(\xb)$, and our goal is to learn the (nonlinear) solution operator mapping from the permeability field $b$ to the corresponding solution field $p$.

\begin{table*}[h!]
\caption{{Test errors and the number of trainable parameters in solution operator learning.
}}
\label{tab:Darcy_results}
\begin{center}
{\small    \centering
\begin{tabular}{cc|r|c|c}
\hline
Case & model & \#param &Linear Operator: {$g\rightarrow p$} &Nonlinear Operator: {$b\rightarrow p$}\\
\cline{4-5}
\hline
$d$ = 20, $d_k$=20& Discrete-NAO & 161991  & 8.61\% & {\bf 10.84\%}\\
$900$ samples&NAO & 89778 & {\bf 8.33\%} & 11.40\% \\
\hline
$d$ = 50, $d_k$=40&Discrete-NAO & 662163  & 3.28\% & 5.61\% \\
$9000$ samples&NAO & 189234 & {\bf 3.19\%} & {\bf 5.28\%} \\
\hline
\end{tabular}}
\end{center}
\vskip -0.1in
\end{table*}

\begin{figure*}
    \centering
    \includegraphics[width=0.3\textwidth]{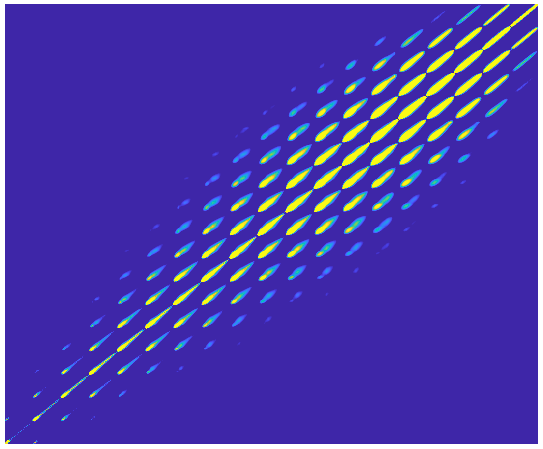}
    \hspace{0.01\textwidth}
    \includegraphics[width=0.3\textwidth]{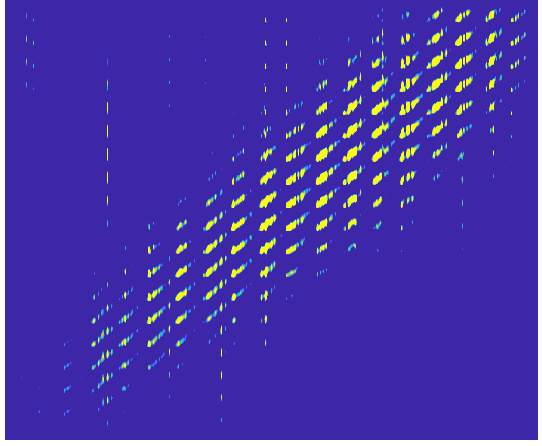}
    \hspace{0.012\textwidth}
    \includegraphics[width=0.299\textwidth]{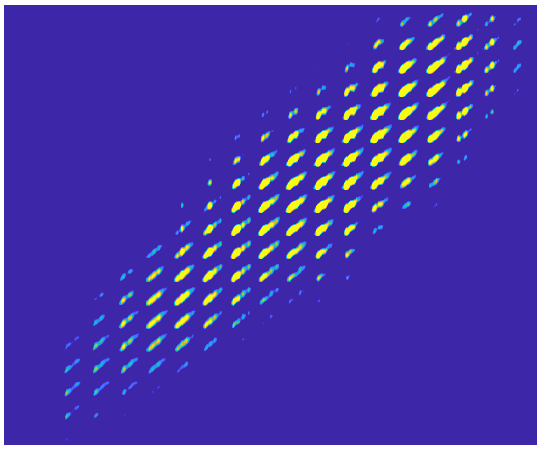}
    \caption{Kernel visualization in experiment 2, where the kernels correspond to the inverse of stiffness matrix: ground truth (left), test kernel from Discrete-NAO (middle), test kernel from NAO (right).}
    \label{fig:stiffness}
\end{figure*}

We report the operator learning results in Table \ref{tab:Darcy_results}, where NAO slightly outperforms Discrete-NAO in most cases, using only 1/2 or 1/3 the number of trainable parameters. On the other hand, we also verify the kernel learning results by comparing the learned kernels in a test case with the ground-truth inverse of stiffness matrix in Figure \ref{fig:stiffness}. Although both Discrete-NAO and NAO capture the major pattern, the kernel from Discrete-NAO again shows a spurious oscillation in regions where the ground-truth kernel has zero value. 
On the other hand, by exploring the kernel map in the integrated knowledge space, the learned kernel from NAO does not exhibit such spurious modes.

\begin{figure}[bt!]
\centering
\includegraphics[width=0.31\textwidth]{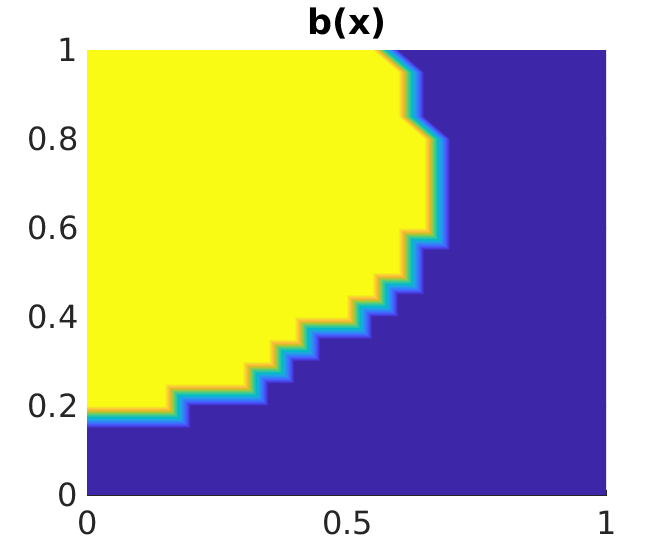}
\hspace{0.008\textwidth}
\includegraphics[width=0.31\textwidth]{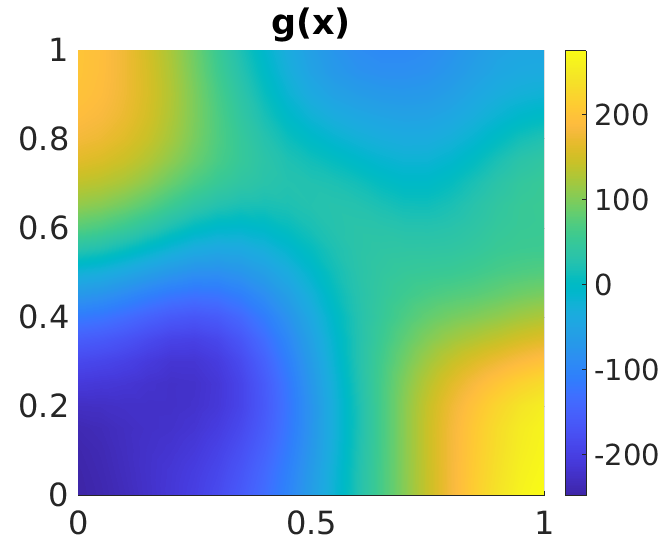}
\hspace{0.008\textwidth}
\includegraphics[width=0.31\textwidth]{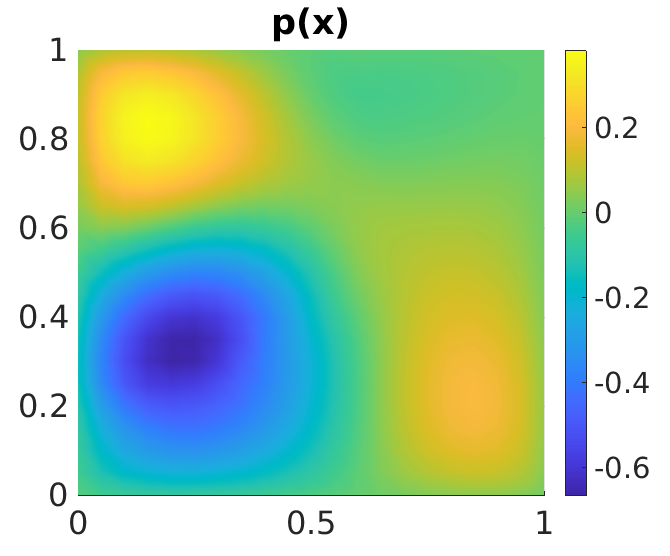}
\includegraphics[width=0.31\textwidth]{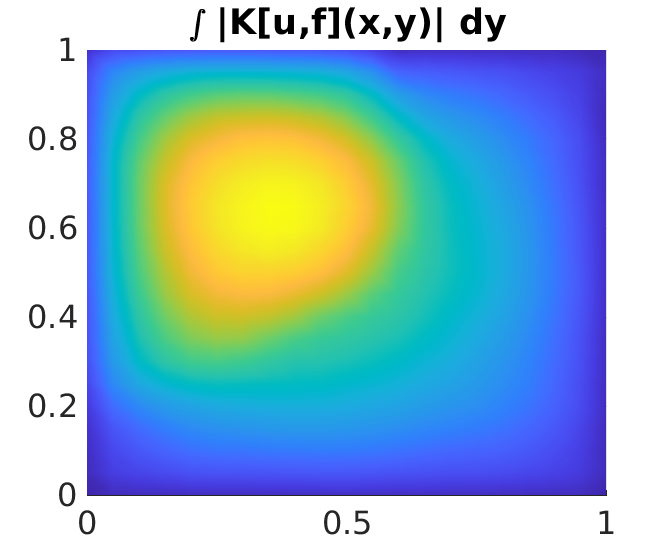}
\hspace{0.018\textwidth}
\includegraphics[width=0.31\textwidth]{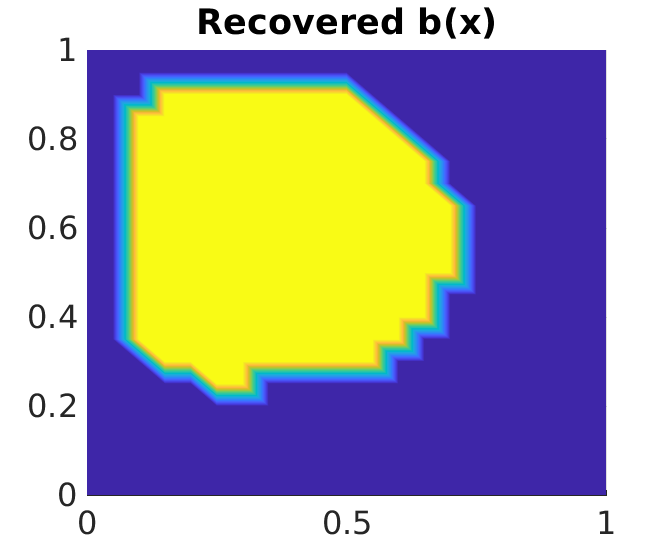}
\caption{Demonstration of the generated data and the recovered microstructure from the learned kernel in Example 2. Top row: the ground-truth two-phase material microstructure from a test task (left), an exemplar loading field instance (middle), and the corresponding solution field instance (right). 
Bottom row: summation of the learned kernel for each line, corresponding to the total interaction of all material points (left), and the discovered two-phase material microstructure after thresholding (right). Note that Dirichlet boundary conditions are applied to all the samples. As a result, the measurement pairs $(p(x),g(x))$ contain no information near the domain boundary $\partial\Omega$, making it impossible to identify the kernel from data on domain boundaries.}
\label{fig:kernel_eg3_setting2}
\end{figure}

To demonstrate the physical interpretability of the learned kernel, in the first row of Figure \ref{fig:kernel_eg3_setting2} we show the ground-truth microstructure $b(\xb)$, a test loading field instance $g(\xb)$, and the corresponding solution $p(\xb)$. By taking the summation of the kernel strength on each row, one can discover the interaction strength of each material point $x$ with its neighbors. As this strength is related to the permeability field $b(x)$, the underlying microstructure can be recovered accordingly. In the bottom row of Figure \ref{fig:kernel_eg3_setting2}, we demonstrate the discovered microstructure of this test task. We note that the discovered microstructure is smoothed out due to the continuous setting of our learned kernel (as shown in the bottom left plot), and a thresholding step is performed to discover the two-phase microstructure. The discovered microstructure (bottom right plot) matches well with the hidden ground-truth microstructure (left plot), except for regions near the domain boundary. This mismatch is due to the applied Dirichlet-type boundary condition ($p(x)=0$ on $\partial\Omega$) in all samples, which leads to the measurement pairs $(p(x),g(x))$ containing no information near the domain boundary $\partial\Omega$ and makes it impossible to identify the kernel on boundaries.


\subsection{Heterogeneous material learning}\label{sec:hetero}

In this example, we investigate the learning of heterogeneous and nonlinear material responses using the Mechanical MNIST benchmark \citep{lejeune2020mechanical}. For training and testing, we take 500 heterogeneous material specimens, where each specimen is governed by a Neo-Hookean material with a varying modulus converted from the MNIST bitmap images. On each specimen, 200 loading/response data pairs are provided. 
Two generalization scenarios are considered. 1) We mix the data from all numbers and randomly take $10\%$ of specimens for testing. This scenario corresponds to an ID test. 2) We leave all specimens corresponding to the number `9' for testing, and use the rest for training. This scenario corresponds to an OOD test. The corresponding results are listed in Table \ref{tab:MMNIST_results}, where NAO again outperforms its discrete counterpart, especially in the small-data regime and OOD setting.

\begin{table*}[h!]
\caption{{Test errors and the number of trainable parameters in heterogeneous material learning.
}}
\label{tab:MMNIST_results}
\begin{center}
{\small    \centering
\begin{tabular}{cc|r|cc}
\hline
Case & model & \#param &ID test & OOD test\\
\hline
$d$ = 40, $d_k$=40& Discrete-NAO & 5,469,528  & 7.21\% & 7.95\% \\
$22500$ samples&NAO &  142,534 & {\bf 6.57\%} &{\bf 6.26\%}\\
\hline
$d$ = 100, $d_k$=100&Discrete-NAO & 7,353,768 & 6.34\% & 6.01\% \\
$45000$ samples&NAO & 303,814 & {\bf 4.75\%}& {\bf 5.58\%}\\
\hline
\end{tabular}}
\end{center}
\vskip -0.15in
\end{table*}



\section{Conclusion}


We propose Nonlocal Attention Operator (NAO), a novel NO architecture to simultaneously learn both the forward (modeling) and inverse (discovery) solvers in physical systems from data. In particular, NAO learns the function-to-function mapping based on an integral NO architecture and provides a surrogate forward solution predictor. In the meantime, the attention mechanism is crafted in building a kernel map from input-output function pairs to the system's function parameters, offering zero-shot generalizability to new and unseen physical systems. As such, the kernel map explores in the function space of identifiability, resolving the enduring ill-posedness in inverse PDE problems. In our empirical demonstrations, NAO outperforms all selected baselines on multiple datasets of inverse PDE problems and out-of-distribution generalizability tasks. 
\newline
{\bf Broader Impacts: } Beyond its merits in forward/inverse PDE modeling, our work represents an initial exploration in understanding the attention mechanism in physics modeling, and paves a theoretical path towards building a foundation model in scientific ML.
\newline
{\bf Limitations:} Due to limited computational resource, our experiments focus on learning from a small to medium number ($<500$) of similar physical systems. It would be beneficial to expand the coverage and enable learning across different types of physical systems.


\begin{ack}
S.~Jafarzadeh would like to acknowledge support by the AFOSR grant FA9550-22-1-0197, and Y.~Yu would like to acknowledge support by the National Science Foundation (NSF) under award DMS-1753031. Portions of this research were conducted on Lehigh University's Research Computing infrastructure partially supported by NSF Award 2019035. F.~Lu would like to acknowledge support by NSF DMS-2238486. 

This article has been authored by an employee of National Technology and Engineering Solutions of Sandia, LLC under Contract No. DE-NA0003525 with the U.S. Department of Energy (DOE). The employee owns all right, title and interest in and to the article and is solely responsible for its contents. The United States Government retains and the publisher, by accepting the article for publication, acknowledges that the United States Government retains a non-exclusive, paid-up, irrevocable, worldwide license to publish or reproduce the published form of this article or allow others to do so, for United States Government purposes. The DOE will provide public access to these results of federally sponsored research in accordance with the DOE Public Access Plan https://www.energy.gov/downloads/doe-public-access-plan.
\end{ack}


\newpage
\appendix

\section{Riemann sum approximation derivations}\label{sec:Riemann}

In this section, we show that the discrete attention operator \eqref{eq:K-attn_Layer} can be seen as the Riemann sum approximation of the nonlocal attention operator in \eqref{eqn:metamodel_full}, in the continuous limit. Without loss of generality, we consider a uniform discretization with grid size $\Delta x$. Denoting the $l-$th layer output as $\Ub^{(l)}=\left(g^{(l)}_j(x_k)\right)_{1\leq j\leq d,1\leq k\leq N}$, $\Fb^{(l)}=\left(f^{(l)}_j(x_k)\right)_{1\leq j\leq d,1\leq k\leq N}$, and $\dfrac{1}{\sqrt{d_k}}\Wb_l^{Q}\Wb_l^{K}$ as $\Wb_l^{QK}\in\real^{d\times d}$, for the $l-$th ($l<L$) layer update in \eqref{eq:K-attn_Layer} writes:
{\small\begin{align*}
\Xb^{(l)}[\alpha,\beta]=&\sum_{\gamma=1}^{2N}\sum_{\omega,\nu=1}^{d}\sum_{\lambda=1}^{d_k}\left(\frac{1}{\sqrt{d_k}} \Xb^{(l-1)}[\alpha,\omega] \Wb_{l}^Q[\omega,\lambda] \Wb_{l}^K[\nu,\lambda] \Xb^{(l-1)}[\gamma,\nu]\right)\Xb^{(l-1)}[\gamma,\beta]+\Xb^{(l-1)}[\alpha,\beta]\\
=&\sum_{\gamma=1}^{2N}\sum_{\omega,\nu=1}^{d}\left( \Xb^{(l-1)}[\alpha,\omega] \Wb_{l}^{QK}[\omega,\nu] \Xb^{(l-1)}[\gamma,\nu]\right)\Xb^{(l-1)}[\gamma,\beta]+\Xb^{(l-1)}[\alpha,\beta].
\end{align*}}
That means, $d_k$ only controls the rank bound when lifting each point-wise feature via $W_l^{QK}$, while $W_l^{QK}$ characterizes the (trainable) interaction of $d$ input function pair instances. Moreover, when $\alpha\leq N$, $\Xb^{(l)}[\alpha,\beta]=g^{(l)}_\beta(x_\alpha)$. When $\alpha>N$, $\Xb^{(l)}[\alpha,\beta]=f^{(l)}_\beta(x_{\alpha-N})$. Then,
{\small\begin{align*}
g^{(l)}_\beta(x_\alpha)=&\sum_{\gamma=1}^{N}\sum_{\omega,\nu=1}^{d}\left( \Xb^{(l-1)}[\alpha,\omega] \Wb_{l}^{QK}[\omega,\nu] \Xb^{(l-1)}[\gamma,\nu]\right)\Xb^{(l-1)}[\gamma,\beta]\\
&+\sum_{\gamma=N+1}^{2N}\sum_{\omega,\nu=1}^{d}\left(
\Xb^{(l-1)}[\alpha,\omega] \Wb_{l}^{QK}[\omega,\nu] \Xb^{(l-1)}[\gamma,\nu]\right)\Xb^{(l-1)}[\gamma,\beta]+g^{(l-1)}_\beta(x_\alpha)\\
=&\sum_{\gamma=1}^{N}\sum_{\omega,\nu=1}^{d}\left( g^{(l-1)}_\omega(x_\alpha) \Wb_{l}^{QK}[\omega,\nu] g^{(l-1)}_\nu(x_\gamma)\right)g^{(l-1)}_\beta(x_\gamma)\\
&+\sum_{\tilde{\gamma}=1}^{N}\sum_{\omega,\nu=1}^{d}\left( g^{(l-1)}_\omega(x_\alpha) \Wb_{l}^{QK}[\omega,\nu] f^{(l-1)}_\nu(x_{\tilde{\gamma}})\right)f^{(l-1)}_\beta(x_{\tilde{\gamma}})+g^{(l-1)}_\beta(x_\alpha).
\end{align*}}
Similarly,
{\small\begin{align*}
f^{(l)}_\beta(x_\alpha)=&\sum_{\gamma=1}^{N}\sum_{\omega,\nu=1}^{d}\left( \Xb^{(l-1)}[\alpha,\omega] \Wb_{l}^{QK}[\omega,\nu] \Xb^{(l-1)}[\gamma,\nu]\right)\Xb^{(l-1)}[\gamma,\beta]\\
&+\sum_{\gamma=N+1}^{2N}\sum_{\omega,\nu=1}^{d}\left( \Xb^{(l-1)}[\alpha,\omega] \Wb_{l}^{QK}[\omega,\nu] \Xb^{(l-1)}[\gamma,\nu]\right)\Xb^{(l-1)}[\gamma,\beta]+f^{(l-1)}_\beta(x_\alpha)\\
=&\sum_{\gamma=1}^{N}\sum_{\omega,\nu=1}^{d}\left( f^{(l-1)}_\omega(x_\alpha) \Wb_{l}^{QK}[\omega,\nu] g^{(l-1)}_\nu(x_\gamma)\right)g^{(l-1)}_\beta(x_\gamma)\\
&+\sum_{\tilde{\gamma}=1}^{N}\sum_{\omega,\nu=1}^{d}\left( f^{(l-1)}_\omega(x_\alpha) \Wb_{l}^{QK}[\omega,\nu] f^{(l-1)}_\nu(x_{\tilde{\gamma}})\right)f^{(l-1)}_\beta(x_{\tilde{\gamma}})+f^{(l-1)}_\beta(x_\alpha).
\end{align*}}
With the Riemann sum approximation: $\int_\omg p(x) dx \approx \Delta x^D \sum_{k=1}^{N_x}p(x_k)$, one can further reformulate above derivations as:
{\small\begin{align*}
g^{(l)}_\beta(x)\approx&\dfrac{1}{\Delta x^D}\int_{\omg}\sum_{\omega,\nu=1}^{d}\left( g^{(l-1)}_\omega(x) \Wb_{l}^{QK}[\omega,\nu] g^{(l-1)}_\nu(y)\right)g^{(l-1)}_\beta(y)dy\\
&+\dfrac{1}{\Delta x^D}\int_{\omg}\sum_{\omega,\nu=1}^{d}\left( g^{(l-1)}_\omega(x) \Wb_{l}^{QK}[\omega,\nu] f^{(l-1)}_\nu(y)\right)f^{(l-1)}_\beta(y)dy+g^{(l-1)}_\beta(x),
\end{align*}}
{\small\begin{align*}
f^{(l)}_\beta(x)\approx &\dfrac{1}{\Delta x^D}\int_{\omg}\sum_{\omega,\nu=1}^{d}\left( f^{(l-1)}_\omega(x) \Wb_{l}^{QK}[\omega,\nu] g^{(l-1)}_\nu(y)\right)g^{(l-1)}_\beta(y)dy\\
&+\dfrac{1}{\Delta x^D}\int_{\omg}\sum_{\omega,\nu=1}^{d}\left( f^{(l-1)}_\omega(x) \Wb_{l}^{QK}[\omega,\nu] f^{(l-1)}_\nu(y)\right)f^{(l-1)}_\beta(y)dy+f^{(l-1)}_\beta(x).
\end{align*}}
Therefore, the attention mechanism of each layer is in fact an integral operator after a rescaling:
\begin{equation}
\left(\begin{array}{c}
g^{(l)}(x)\\
f^{(l)}(x)\\
\end{array}\right)=\int_\omg K^{(l)}(x,y)
\left(\begin{array}{c}
g^{(l-1)}(y)\\
f^{(l-1)}(y)\\
\end{array}\right)dy+
\left(\begin{array}{c}
g^{(l-1)}(x)\\
f^{(l-1)}(x)\\
\end{array}\right),
\end{equation}
with the kernel $K^{(l)}(x,y)$ defined as:
{\small\begin{equation}
\left[\begin{array}{cc}
\sum_{\omega,\nu=1}^{d}\left( g^{(l-1)}_\omega(x) \Wb_{l}^{QK}[\omega,\nu] g^{(l-1)}_\nu(y)\right)&\sum_{\omega,\nu=1}^{d}\left( g^{(l-1)}_\omega(x) \Wb_{l}^{QK}[\omega,\nu] f^{(l-1)}_\nu(y)\right)\\
\sum_{\omega,\nu=1}^{d}\left( f^{(l-1)}_\omega(x) \Wb_{l}^{QK}[\omega,\nu] g^{(l-1)}_\nu(y)\right)&\sum_{\omega,\nu=1}^{d}\left( f^{(l-1)}_\omega(x) \Wb_{l}^{QK}[\omega,\nu] f^{(l-1)}_\nu(y)\right)\\
\end{array}\right].
\end{equation}}
For the $L-$th layer update, we denote the approximated value of $f_\beta(x_\alpha)$ as $\tilde{f}_\beta(x_\alpha):=\Xb_{\text{out}}[\alpha,\beta]$, then
{\small\begin{align*}
\tilde{f}_\beta(x_\alpha)=&\sum_{\gamma=1}^N K[\mathbf{u}_{1:d},\mathbf{f}_{1:d};\theta][\alpha,\gamma] g[u_\beta](x_\gamma)\\
=&\sum_{\lambda,\gamma=1}^{N}\sum_{\omega,\nu=1}^{d} W^{P,u}[\alpha,\lambda] \left(\Ub^{(L-1)}[\lambda,\omega] W_L^{QK}[\omega,\nu] \Ub^{(L-1)}[\gamma,\nu]  \right)  g[u_\beta](x_\gamma)\\
&+ \sum_{\lambda,\gamma=1}^{N}\sum_{\omega,\nu=1}^{d} W^{P,f}[\alpha,\lambda]\left(\Fb^{(L-1)} [\lambda,\omega] W_L^{QK}[\omega,\nu] \Ub^{(L-1)}[\gamma,\nu] \right)g[u_\beta](x_\gamma)\\
=&\sum_{\lambda,\gamma=1}^{N}\sum_{\omega,\nu=1}^{d} W^{P,u}[\alpha,\lambda] \left(g^{(L-1)}_{\omega}(x_\lambda) W_L^{QK}[\omega,\nu] g^{(L-1)}_{\nu}(x_\gamma)  \right)  g[u_\beta](x_\gamma)\\
&+ \sum_{\lambda,\gamma=1}^{N}\sum_{\omega,\nu=1}^{d} W^{P,f}[\alpha,\lambda]\left(f^{(L-1)}_{\omega}(x_\lambda) W_L^{QK}[\omega,\nu] g^{(L-1)}_{\nu}(x_\gamma) \right)g[u_\beta](x_\gamma)\\
\approx &\dfrac{1}{\Delta x^{2D}}\int_{\omg}\int_{\omg}\sum_{\omega,\nu=1}^{d} W^{P,u}(x_\alpha,z) \left(g^{(L-1)}_{\omega}(z) W_L^{QK}[\omega,\nu] g^{(L-1)}_{\nu}(y)  \right)dz\;  g[u_\beta](y)dy\\
&+ \dfrac{1}{\Delta x^{2D}}\int_{\omg}\int_{\omg}\sum_{\omega,\nu=1}^{d} W^{P,f}(x_\alpha,z) \left(f^{(L-1)}_{\omega}(z) W_L^{QK}[\omega,\nu] g^{(L-1)}_{\nu}(y)  \right)dz\;  g[u_\beta](y)dy.
\end{align*}}
Hence, a (rescaled) continuous limit of the kernel writes:
\begin{align*}
K[\mathbf{u}_{1:d},\mathbf{f}_{1:d};\theta](x,y)=&\int_{\omg}\sum_{\omega,\nu=1}^{d} W^{P,u}(x,z) \left(g^{(L-1)}_{\omega}(z) W_L^{QK}[\omega,\nu] g^{(L-1)}_{\nu}(y)  \right)dz\\
&+\int_{\omg}\sum_{\omega,\nu=1}^{d} W^{P,f}(x,z) \left(f^{(L-1)}_{\omega}(z) W_L^{QK}[\omega,\nu] g^{(L-1)}_{\nu}(y)  \right)dz.
\end{align*}

\section{Proofs and connection with regularized estimator}

\subsection{Proofs}\label{sec:appd-proofs}

\begin{proof}[Proof of Lemma \ref{lemma:attn-limit}] With $\{r_k\}_{k=1}^N$, we can write the attention-based kernel in \eqref{eq:kernel_map} as 
\begin{equation}\label{eq:attn_kernel_val}
\begin{aligned}
K[\mathbf{u}_{1:d},\mathbf{f}_{1:d};\theta](r_k) = & \sum_{k'=1}^N  W^{P,u}(r_{k'}) \sigma \left( \sum_{l=1}^{d_k} \sum_{j=1}^d \sum_{j=1}^d \mathbf{u}_{j}(r_{k'}) W^Q[j,l] \cdot W^K[j',l] \mathbf{u}_{j'}(r_k)  \right)  \\
 &+W^{P,f}\sigma\left(\sum_{l=1}^{d_k}  \sum_{j=1}^d \sum_{j'=1}^d \left[\mathbf{f}_{j} W^Q[j,l]\cdot W^K[j',l] \mathbf{u}_{j'}(r_k) \right]\right).  
 \end{aligned}
 \end{equation}
 
Denoting $W^{QK}[j',j] =\sum_{l=1}^{d_k} W^Q[j,l]\cdot W^K[j',l]  $, we write the kernel in \eqref{eq:attn_kernel_val} as 
\begin{align*}
K[\mathbf{u}_{1:d},\mathbf{f}_{1:d};\theta](r)=&
\sum_{r'=1}^N W^{P,u}(|r'|)\sigma\left(\sum_{j=1}^d \sum_{j'=1}^d \left[g[u](r',x_{i_j}) W^{QK}[j,j'] g[u](r,x_{i_{j'}})  \right]\right)  
\\
&+W^{P,f}\sigma\left(\sum_{i=1}^d \sum_{j=1}^d \left[f(x_{i_{j'}})W^{QK}[j,j']  g[u](r,x_{i_j})\right]\right).
\end{align*}
Then, as $d\to \infty$ is achieved by refining the spatial mesh, viewing the summation in $j$ as Riemann sum, 
\[
\lim_{d\to \infty}\sum_{j=1}^d \sum_{j'=1}^d  g[u](r',x_{i_j}) W^{QK}[j,j'] g[u](r,x_{i_{j'}}) =\int \int g[u](r,x) W^{QK}(x,y)  g[u](s,y) dxdy, 
\]
where the integral exists since $g[u](r,x) W^{QK}(x,y)$ is bounded. Sending also the number of tokens, $N$, to infinity, we obtain the limit attention model in \eqref{eq:att_limit}. 
\end{proof}

\begin{proof}[Proof of Lemma \ref{lemma:ID}]
The proof is adapted from \cite{lu2023nonparametric,lu2022data}. Write $K(r) = K[\mathbf{u}^\eta_{1:d},\mathbf{f}^\eta_{1:d}](r)$. Notice that the loss function in \eqref{eq:lossFn} can be expanded as 
\begin{align*}
\mcE(K) = &\dfrac{1}{n_{train}} \sum_{\eta=1}^{n_{train}} \sum_{i=1}^{d_u} \int_{\omg} \left[\int_{0}^{\delta} K(r)  g[u_i^\eta](r,x))dr-f_i^\eta(x)\right]^2 dx\\
=&\dfrac{1}{n_{train}} \sum_{\eta=1}^{n_{train}} \sum_{i=1}^{d_u} \int_{0}^{\delta}\int_{0}^{\delta}  K(s) K(r) \int_{\omg} g[u_i^\eta](s,x)) g[u_i^\eta](r,x)) dx dr ds\\
&-\dfrac{2}{n_{train}} \sum_{\eta=1}^{n_{train}} \sum_{i=1}^{d_u}  \int_{0}^{\delta} K(r) \int_{\omg} g[u_i^\eta](r,x) f_i^\eta(x) dx dr +Const.\\
= &  \langle  \mcL_{\bar{G}}  K, K \rangle_{L^2_\rho}- 2 \langle  K, K^D \rangle_{L^2_\rho}+Const., 
\end{align*}
where $\mcL_{\bar{G}}: L^2_\rho\to L^2_\rho$ is the integral operator 
\[
\mcL_{\bar{G}} K(s) := \int_0^\delta K(r)\bar{G}(r,s)dr\]
and $K^D$ is the Riesz representation of the bounded linear functional 
\[
\langle  K, K^D \rangle_{L^2_\rho} = \dfrac{1}{n_{train}} \sum_{\eta=1}^{n_{train}} \sum_{i=1}^{d_u}  \int_{0}^{\delta} K(r) \int_{\omg} g[u_i^\eta](r,x) f_i^\eta(x) dx dr. 
\]
Thus, the quadratic loss function has a unique minimizer in $\mathrm{Null}(\mcL_{\bar{G}})^\perp$. 

Meanwhile, since the data pairs are continuous with compact support, the function $\bar{G}$ is a square-integrable reproducing kernel. Thus, the operator $\mcL_{\bar{G}}$ is compact and $H_G= \mcL_{\bar{G}}^{1/2}L^2_\rho$. 
Then, $\mathrm{Null}(\mcL_{\bar{G}})^\perp = \overline{H_G} $, where the closure is with respect to $L^2_\rho$. 
\end{proof}


\subsection{Connection with regularized estimator}\label{sec:appd-reguEst}
Consider the inverse problem of estimating the nonlocal kernel $K$ given a data pair $(u,f)$. In the classical variational approach, one seeks the minimizer of the following loss function  
\begin{align}\label{eq:E-K}
\mathcal{E}(K)= & \int_{\omg} \left[\int_{0}^{\delta} K(r) g[u](r,x)dr-f(x)\right]^2 dx
\end{align}
The inverse problem is ill-posed in the sense that the minimizer can often be non-unique or sensitive to the noise or measurement error in data $(u,f)$. 
Thus, regularization is crucial to produce a stable solution. 

To connect with the attention-based model, we consider regularizing using an RKHS $H_W$ with a square-integrable reproducing kernel $W$. One seeks an estimator in $H_W$ by regularizing the loss with the $\|K\|_{H_W}^2$, and minimizes the regularized loss function  
\begin{align}\label{eq:E_Wregu}
\mathcal{E}_{\lambda,W}(K)= &\mathcal{E}(K)  +\lambda \int_{0}^{\infty}\int_{0}^{\infty}  K(s) K(r)  W(r,s) dr ds. 
\end{align}

The next lemma shows that the regularized estimator is a nonlinear function of the data pair $(u,f)$, where the nonlinearity comes from the kernel $G_u$ and the parameter $\lambda_*$. 
\begin{lemma}\label{lemma:regu_estK}
The regularized loss function in $\mathcal{E}_{\lambda,W}(K)$ in \eqref{eq:E_Wregu} is
\begin{align}\label{eq:E_Wregu2}
\mathcal{E}_{\lambda,W}(K)=&  \int_0^\infty \int_0^\infty K(r)K(s) [ G_u(r,s) + \lambda W(r,s) ]drds - 2 \int_0^\infty K(r)K^{u,f}(r)dr + Const.,
\end{align}
where $G_u$ is defined in \eqref{eq:Gu}. 
Its minimizer is
\begin{equation}\label{eq:regu_estK}
\widehat{K} = ( \mcL_{G_u} + \lambda_* \mcL_W)^{-1} K^{u,f}, 
\end{equation}
where $\mcL_{G_u}$ and $\mcL_W$ are integral operators with integral kernels $G_u$ defined in \eqref{eq:Gu} and $W$, 
\[
\mcL_{G_u}K(s) := \int_0^\infty K(r)G_{u}(r,s)dr, \quad \mcL_{W}K(s) := \int_0^\infty K(r)W(r,s)dr,
\]
 $\lambda_*$ is the optimal hyper-parameter controlling the strength of regularization, and $K^{u,f}(r)= \int_{\omg}g[u](r,x) f(x) dx$ is a function determined by the data $(u,f)$.  
\end{lemma}

When there is no prior information on the regularization, which happens often for the learning of the kernel, one can use the data-adaptive RKHS $H_{G_u}$ with the reproducing kernel $G_u$ determined by data:
 \begin{equation}\label{eq:Gu}
 \begin{aligned}
 G_u(r,s)
= &   \int_{\omg} g[u](r,x) g[u](s,x) dx. 
\end{aligned}
 \end{equation}
 \cite{lu2023nonparametric} shows that this regularizer can lead to consistent convergent estimators. 

\begin{remark}[Discrete data and discrete inverse problem] In practice, the datasets are discrete. One can view the discrete inverse problem as a discretization of the continuous inverse problem. Assume that the integrands are compactly supported and when the integrals are approximated by Riemann sums, we can write loss function for discrete $\mathbf{K} = ( K(r_1), \ldots, K(r_N))^\top \in \real^{N\times 1}$ as  
\begin{align}\label{eq:E_Wregu-disc}
\mathcal{E}_{\lambda,W}(K)\approx &  \sum_{k,k'=1}^{N,N} K(r_k)K(r_{k'}) [ G_u(r_k,r_{k'}) + \lambda W(r_k,r_{k'}) ] (\Delta r)^2 - 2 \sum_{k=1}^N K(r_k)K^{u,f}(r_k) \Delta r  + Const.. \notag\\ 
=& \mathbf{K} ^\top [ \mathbf{G}_u + \lambda \mathbf{W}] \mathbf{K} - 2 \mathbf{K}^\top \mathbf{K}^{u,f} + Const., 
\end{align}
where, recalling the definition of the token $\mathbf{u}_{1:d} $ in \eqref{eq:u_to_token}, 
\[
\mathbf{G}_u = ( G_u(r_k,r_{k'}) )_{1\leq k,k'\leq K} = ( \int_{\omg} g[u](x,r_k) g[u](x,r_{k'}) dx)_{1\leq k,k'\leq K} \approx \mathbf{u}_{1:d}  \mathbf{u}_{1:d}^\top,    
\]
 $\mathbf{W} =  \left( W(r_k,r_{k'}) \right)_{1\leq k,k'\leq K}$ and $\mathbf{K}^{u,f} = \left( \int_{\omg}g[u](r_k,x) f(x) dx \right)_{1\leq k\leq N}$. 

The minimizer of this discrete loss function with the optimal hyper-parameter $\lambda_*$ is 
\[
\widehat{\mathbf{K}} = (\mathbf{G}_u + \lambda_* \mathbf{W} )^{-1} \mathbf{K}^{u,f}. 
\]
In particular, when taking $\mathbf{W} = \mathbf{G}_u$ and using the Neumann series $(\lambda_*^{-1}\mathbf{G}_u^2 +  I )^{-1}= \sum_{k=0}^\infty (-1)^k \lambda_*^{-k}\mathbf{G}_u^{2k}$, we have 
\[
\widehat{\mathbf{K}} = (\lambda_*^{-1}\mathbf{G}_u^2 +  I )^{-1} \lambda_*^{-1}\mathbf{G}_u\mathbf{K}^{u,f}= \lambda_*^{-1}\mathbf{G}_u\mathbf{K}^{u,f}  
- \lambda_*^{-2}\mathbf{G}_u^{3} \mathbf{K}^{u,f} + \sum_{k=2}^\infty (-1)^k \lambda_*^{-k}\mathbf{G}_u^{2k} \mathbf{G}_u \mathbf{K}^{u,f} . 
\]
In particular, $\lambda_*$ depends on both $\mathbf{G}_u $ and $\mathbf{K}^{u,f} $. Hence, the estimator $\widehat{\mathbf{K}}$ is nonlinear in $\mathbf{G}_u$ and $\mathbf{K}^{u,f} $, and it is important to make the attention depend nonlinearly on the token $\mathbf{u}_{1:d}$, as in \eqref{eq:kernel_map}. 
\end{remark}
\begin{proof}[Proof of Lemma \ref{lemma:regu_estK}] Since $K$ is radial and noticing that 
$$
\int_{|\xi|=1} (u(x+r\xi)-u(x)) d\xi = u(x+r)+ u(x- r)-2u(x) = g[u](r,x) 
$$ 
since $\xi\in \real^1$,  we can write
\begin{align*}
& \int_{-\infty}^{\infty}\int_{-\infty}^{\infty}  K(|s|) K(|r|) \int_{\omg} (u(x+s)-u(x))  (u(x+r)-u(x))dx dr ds \\
= & \int_0^\infty \int_0^\infty K(r)K(s) \int_{|\xi'|=1}\int_{|\xi|=1} \int_{\omg}(u(x+s\xi')-u(x))  (u(x+r\xi)-u(x)) dx d\xi d\xi'  \\
= & \int_0^\infty \int_0^\infty K(r)K(s) \int_{\omg}g[u](r,x)  g[u](s,x)  dx dr ds    \\
= & \int_0^\infty \int_0^\infty K(r)K(s) G_u(r,s) drds. 
\end{align*}
Meanwhile, by the Riesz representation theorem, there exists a function $K^{u,f}\in L^2(0,\infty)$ such that 
\begin{align*}
& \int_{-\infty}^{\infty} K(|r|) \int_{\omg} (u(x+r)-u(x)) f(x) dx dr \\
= &  \int_0^\infty K(r) \int_{|\xi|=1}  \int_{\omg} (u(x+r)-u(x)) f(x) dx d\xi dr \\
= &  \int_0^\infty K(r) \int_{\omg}g[u](r,x) f(x) dx dr 
=  \int_0^\infty K(r)K^{u,f}(r)dr. 
\end{align*}
Combining these two equations, we can write the loss function as
\begin{align*}
\mathcal{E}(K)= & \int_{\omg} \left[\int_{-\infty}^{\infty} K(|s|)  (u(x+s)-u(x))ds-f(x)\right]^2 dx\\
=&\int_{-\infty}^{\infty}\int_{-\infty}^{\infty}  K(|s|) K(|r|) \int_{\omg} (u(x+s)-u(x))  (u(x+r)-u(x))dx dr ds\\
&- \int_{-\infty}^{\infty} K(|s|) \int_{\omg} (u(x+s)-u(x)) f(x) dx ds +Const. \\
= & \int_0^\infty \int_0^\infty K(r)K(s) G_u(r,s) drds - 2 \int_0^\infty K(r)K^{u,f}(r)dr + Const.. 
\end{align*}
Then, we can write the regularized loss function  $\mathcal{E}_{\lambda,W}(K)$ as 
\begin{align*}
\mathcal{E}_{\lambda,W}(K)=&  \int_0^\infty \int_0^\infty K(r)K(s) [ G_u(r,s) + \lambda W(r,s) ]drds - 2 \int_0^\infty K(r)K^{u,f}(r)dr + Const.\\
= & \langle  ( \mcL_{G_u} + \lambda \mcL_W) K, K \rangle_{L^2(0,\infty)} - 2  \langle K,  K^{u,f}\rangle_{L^2(0,\infty)} + Const.
\end{align*}
Selecting the optimal hyper-parameter $\lambda_*$, which depends on both $(u,f)$ and $W$, and setting the Fr\'echet derivative of $\mathcal{E}_{\lambda,W}$ over $L^2(0,\infty)$ to be zero, we obtain the regularized estimator in \eqref{eq:regu_estK}.
\end{proof}

\section{Data generation and additional discussion}\label{sec:exp_more}


\subsection{Example 1: radial kernel learning}\label{sec:appd-ex1}

In all settings except the ``single task'' one, all kernels act on the same set of functions $\{u_i\}_{i=1,2}$ with $u_1= \cos(x)\mathbf{1}_{[-\pi,\pi]}(x)$ and $u_2(x) =\sin(2x)\mathbf{1}_{[-\pi,\pi]}(x)$. In the ``single task'' setting, to create more diverse samples, the single kernel acts on a set of 14 functions: 
$u_{k}= \cos(kx)\mathbf{1}_{[-\pi,\pi]}(x)$, $k=1,\cdots,7$ and $u_{k}(x) =\sin(kx)\mathbf{1}_{[-\pi,\pi]}(x)$, $k=8,\cdots, 14$. In the ground-truth model, the integral $\mcL_{\gamma_\eta}[u_i]$ is computed by the adaptive Gauss-Kronrod quadrature method, which is much more accurate than the Riemann sum integrator that we use in the learning stage.
To create discrete datasets with different resolutions, for each $\Delta x\in0.0125\times\{1,2\}$, we take values $\{u_i,f_i\}_{i=1}^N = \{u_i(x_j),f_i(x_j):x_j \in {[-40,40]}, j=1,\ldots,J\}_{i=1}^N$, where $x_j$ is the grid point on the uniform mesh of size $\Delta x$. We form a training sample of each task by taking $d$ pairs from this task. When taking the token size $d$ and $k$ function pairs, each task contains $\lfloor\frac{2265}{d}\rfloor \times k$ samples.

\subsection{Example 2: solution operator learning}

For one example, we generate the synthetic data based on the Darcy flow in a square domain of size $1\times1$ subjected to Dirichlet boundary conditions. The problem setting is:
$-\nabla(b(x)\nabla p(x))=g(x)$
subjected to $p(x) = 0$ on all boundaries. This equation describes the diffusion in heterogeneous fields, such as the subsurface flow of underground water in porous media. The heterogeneity is represented by the location-dependent conductivity $b(x)$. $p(x)$ is the source term, and the hydraulic height $g(x)$ is the solution. For each data instance, we solve the equation on a $21\times 21$ grid using an in-house finite difference code. We consider 500 random microstructures consisting of two distinct phases. For each microstructure, the square domain is randomly divided into two subdomains with different conductivity of either 12 or 3. Additionally, we consider 100 different $g(x)$ functions obtained via a Gaussian random field generator. For each microstructure, we solve the Darcy problem considering all 100 source terms, resulting in a dataset of $N=10,000$ function pairs in the form of $\{p_i(x_j),g_i(x_j)\}_{i=1}^N$, and $j=1,2,\cdots,441$ where $x_j$'s are the discretization points on the square domain.

In operator learning settings, we note that the permutation of function pairs in each sample should not change the learned kernel, i.e., one should have $K[\mathbf{u}_{1:d},\mathbf{f}_{1:d}]=K[\mathbf{u}_{\sigma(1:d)},\mathbf{f}_{\sigma(1:d)}]$, where $\sigma$ is the permutation operator. Hence, we augment the training dataset by permuting the function pairs in each task. Specifically, with $100$ microstructures (tasks) and $100$ function pairs per task, we randomly permut the function pairs and take $100$ function pairs for $100$ times per task. As a result, we can generate a total of 10000 samples (9000 for training and 1000 for testing) in the form of $\{\ub^\eta_{1:100},\fb^\eta_{1:100}\}_{\eta=1}^{10000}$.

\subsection{Example 3: heterogeneous material learning}

In the Mechanical MNIST dataset, we generate a large set of heterogeneous material responses subjected to mechanical forces. This is similar to the approach in \cite{lejeune2020mechanical}. The material property (stiffness) of the heterogeneous medium is constructed assuming a linear scaling between 1 and 100, according to the gray-scale bitmap of the MNIST images, which results in a set of 2D square domains with properties that vary according to MNIST digit patterns. The problem setting for this data set is the equilibrium equation:$-\nabla \cdot P(x)=f(x)$ subjected to Dirichlet boundary condition of zero displacement and nonzero variable external forces:$f(x)$. $P(x)=\hat P(I + \nabla u(x))$ is the stress tensor and is a nonlinear function of displacement $u(x)$. The choice of material models determines the stress function, and here we employ the Neo-Hookean material model as in \cite{lejeune2020mechanical}. From MNIST, we took 50 samples of each digit resulting in a total of 500 microstructures. We consider 200 different external forces $f(x)$ obtained via a Gaussian random field, and solve the problem for each pair of microstructure and external force, resulting in $N=100,000$ samples. To solve each sample, we use FEniCS finite element package considering a $140\times140$ uniform mesh. We then downsample from the finite element nodes to get values for the solution $u(x)$ and external force $f(x)$ on a the coarser $29\times29$ equi-spaced grid. The resulting dataset is of the form: $\{u_i(x_j),f_i(x_j)\}_{i=1}^N$, and $j=1,2,...,841$ where $x_j$'s are the equi-spaced sampled spatial points.  

Similar to Example 2, we perform the permutation trick to augment the training data. In particular, with $500$ microstructures (tasks) and $200$ function pairs per task, we randomly permute the function pairs and take $100$ function pairs for $100$ times per task. As a result, we can generate a total of 50000 samples, where 45000 are used for training and 5000 for testing.

\section{Computational complexity}\label{sec:appd_complexity}
Denote $(N,d,d_k,L,n_{train})$ as the size of the spatial mesh, the number of data pairs in each training model, the column width of the query/key weight matrices $W^Q$ and $W^K$, the number of layers, and the number of training models. The number of trainable parameters in a discrete NAO is of size $O(L\times d\times d_k+N^2)$. For continuous-kernel NAO, taking a three-layer MLP as a dense net of size $(d, h_1, h_2, 1)$ for the trainable kernels $W^{P,u}$ and $W^{P,f}$ for example, its the corresponding number of trainable parameters is $O(d\times h_1+h_1\times h_2)$. Thus, the total number of trainable parameters for a continuous-kernel NAO is $O(L\times d\times d_k+d\times h_1+h_1\times h_2)$.

The computational complexity of NAO is quadratic in the length of the input and linear in the data size, with $ O([d^2(3N+d_k) + 6N^2d] Ln_{train}) $ flops in each epoch in the optimization. It is computed as follows for each layer and the data of each training model: the attention function takes $ O(d^2d_k+  2Nd^2+4N^2d )$ flops, and the kernel map takes $O(d^2d_k+  Nd^2+2N^2d)$ flops; thus, the total is $O(d^2(3N+d_k) + 6N^2d)$ flops. In inverse PDE problems, we generally have $d\ll N$, and hence the complexity of NAO is $O(N^2d)$ per layer per sample.

Compared with other methods, it is important to note that NAO solves both forward and ill-posed inverse problems using multi-model data. Thus, we don't compare it with methods that solve the problems for a single model data, for instance, the regularized estimator in Appendix B. Methods solving similar problems are the original attention model \citep{vaswani2017attention}, convolution neural network (CNN), and graph neural network (GNN). As discussed in \cite{vaswani2017attention}, these models have a similar computational complexity, if not any higher. In particular, the complexity of the original attention model is $O(N^2d)$, and the complexity of CNN is $O(kNd^2)$ with $k$ being the kernel size, and a full GNN is of complexity $O(N^2d^2)$.

\end{document}